\documentclass[conference]{IEEEtran}
\usepackage{times}

\usepackage[numbers]{natbib}
\usepackage{multicol}
\usepackage[bookmarks=true]{hyperref}
\usepackage{float}

\usepackage{graphicx} 
\usepackage{subcaption}

\usepackage{amsmath}
\DeclareMathAlphabet{\mbf}{OT1}{ptm}{b}{n}

\newcommand{\Real}{\mathbb R}

\newcommand{\mbfhat}[1]{{\hat{\mbf{#1}}}}

\newcommand{\sheepemoji}{%
  \raisebox{-0.2ex}{%
    \includegraphics[height=1.05em]{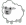}%
  }%
}

\usepackage[dvipsnames]{xcolor}
\usepackage{moreverb,url}
\usepackage{amsmath,amsfonts,amssymb}
\usepackage{tikz,tkz-euclide}
\usepackage{balance}
\usepackage{siunitx}
\usetikzlibrary{decorations,calligraphy}
\usetikzlibrary{calc}
\usetikzlibrary{spy}
\tikzset{>=latex}

\usepackage{booktabs}
\usepackage{tabularx}
\newcolumntype{Y}{>{\centering\arraybackslash}X}
\newcolumntype{L}{>{\arraybackslash}X}

\usepackage{multirow}

\usepackage[nolist,nohyperlinks]{acronym}

\def\vizset{\mathcal{V}}

\def\m{m}

\def\gt{{\text{GT}}}

\def\map{\mathcal{M}}

\def\planar{\mathcal{P}}
\def\edge{\mathcal{E}}

\newcommand\residualplanar[1]{{r_{\planar_j}}}
\newcommand\residualedge[1]{{r_{\edge_j}}}

\def\coordinate{{\mathbf{e}}}

\def\inv{^{\text{-}1}}

\DeclareMathOperator*{\argmin}{arg\,min}
\newcommand{\transpose}{{\ensuremath{\mathsf{T}}}} 
\newcommand{\norm}[1]{\left\Vert#1\right\Vert}

\begin{acronym}[AAAAAAAAA]
    \acro{ate}[ATE]{absolute trajectory error}
    \acro{ba}[BA]{bundle adjustment}
    \acro{drl}[DRL]{direct radar localization}
    \acro{epe}[EPE]{end-pose-error}
    \acro{rmse}[RMSE]{root mean squared error}
    \acro{ndt}[NDT]{normal distribution transform}
    \acro{slam}[SLAM]{simultaneous localization and mapping}
    \acro{icp}[ICP]{iterative closest point}
    \acro{sota}[SOTA]{state-of-the-art}
    \acro{varpro}[VarPro]{variable projection}
\end{acronym}

\begin{document}

\title{Dr-BA\sheepemoji: Separable Optimization for Direct Radar Bundle Adjustment \& Localization}

\author{\authorblockN{Daniil Lisus, Cedric Le Gentil, Timothy D. Barfoot}
\authorblockA{Robotics Institute, University of Toronto, Toronto, Canada\\
Email: \{daniil.lisus; cedric.legentil; tim.barfoot\}@robotics.utias.utoronto.ca\\
}}


%

\maketitle

\begin{abstract}
This paper introduces Dr-BA, a first-of-its-kind radar bundle adjustment (BA) framework that operates directly on 2D spinning radar intensity images.
Unlike camera or lidar sensors, radar is largely unaffected by precipitation, making it a critical modality for autonomous systems that require all-weather robustness.
Existing state estimation approaches using spinning radar typically extract sparse point clouds from range-azimuth-intensity measurements and apply point cloud alignment techniques to estimate vehicle motion, scene structure, or to localize within an existing map.
In contrast, Dr-BA uses the full radar returns from multiple scans to jointly estimate dense maps and sensor poses.
By formulating the problem as a separable optimization, we derive an efficient and general solution that decouples pose estimation from mapping.
In addition to solving the BA problem, this formulation naturally extends to direct radar-only localization (DRL) within a previously built map.
Dr-BA achieves state-of-the-art radar-based BA and cross-session localization performance, demonstrated on more than 200 km of on-road data across five distinct routes. Our implementation is publicly available at \url{https://github.com/utiasASRL/dr_ba}.
\end{abstract}

\IEEEpeerreviewmaketitle

\section{Introduction}
For a robot to function autonomously, it must have an accurate representation of the environment in which it operates.
Consequently, a critical challenge in modern robotic deployment is constructing a reliable map of the environment.
This is typically achieved using \ac{slam} algorithms, in which sensor measurements are used to jointly estimate both the sensor’s trajectory and the surrounding map.
SLAM algorithm performance is commonly evaluated by measuring the accuracy of the estimated sensor trajectory using metrics such as the \ac{ate}.
However, the primary output of a \ac{slam} algorithm is the map itself, and good trajectory accuracy does not necessarily imply high-quality maps for downstream applications.
For instance, a trajectory with zero-mean jitter may yield a low \ac{ate} while producing a locally inconsistent map that is difficult to use in practice.
In this paper, we present Dr-BA, a radar-based \ac{ba} framework designed to produce maps that are both locally and globally consistent.
We evaluate the proposed method through quantitative and qualitative \ac{ba} metrics, and further demonstrate its effectiveness on the downstream task of localization.

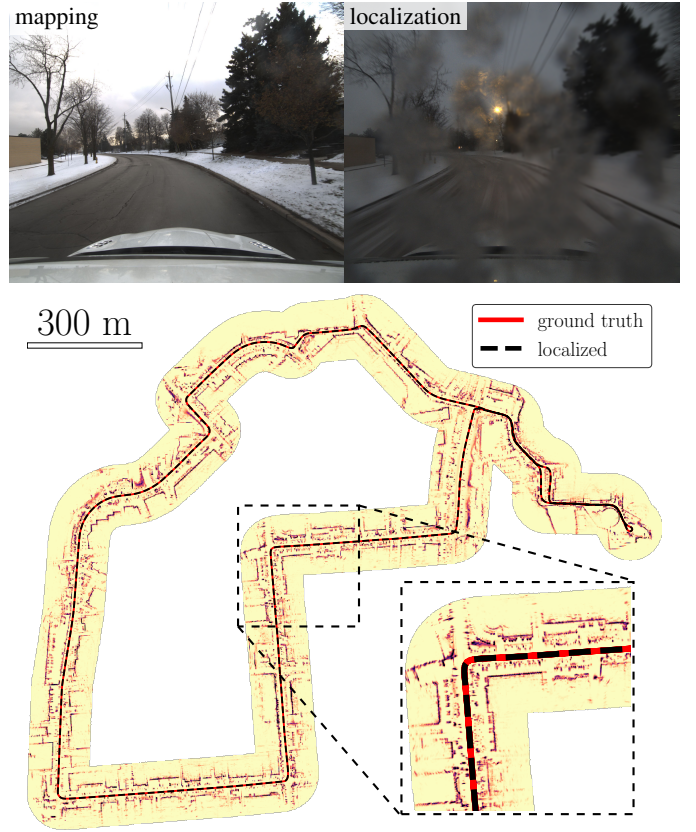
\begin{figure}
    \centering
    \begin{tikzpicture}[spy using outlines={magnification=3, size=3cm, connect spies, dashed, ultra thick}]
        \node[inner sep=0, anchor=north] (map) at (0.25\linewidth,-2.0)
            {\includegraphics[width=\linewidth]{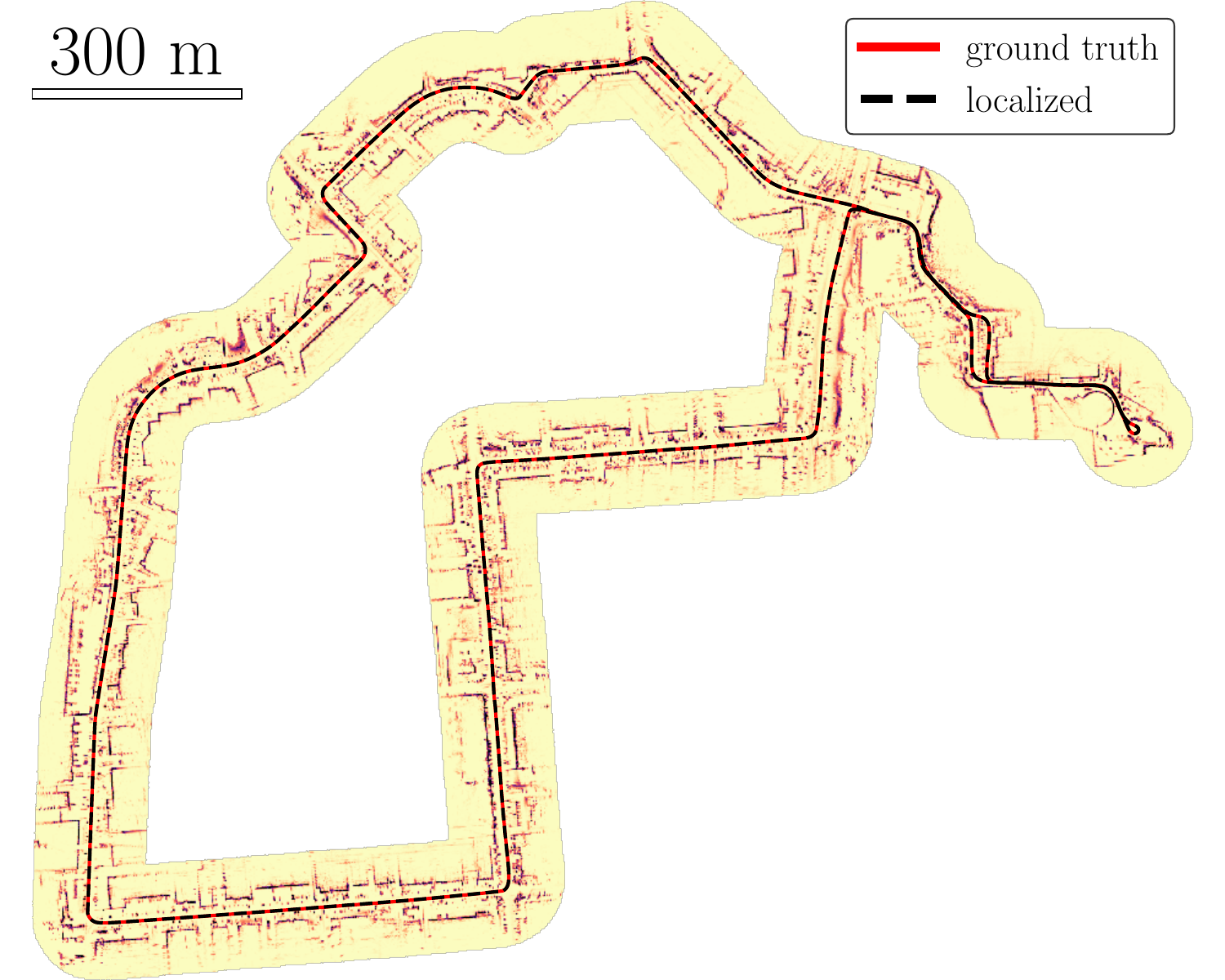}};
        \node[inner sep=0] (clear_img) at (0,0)
            {\includegraphics[width=0.5\linewidth]{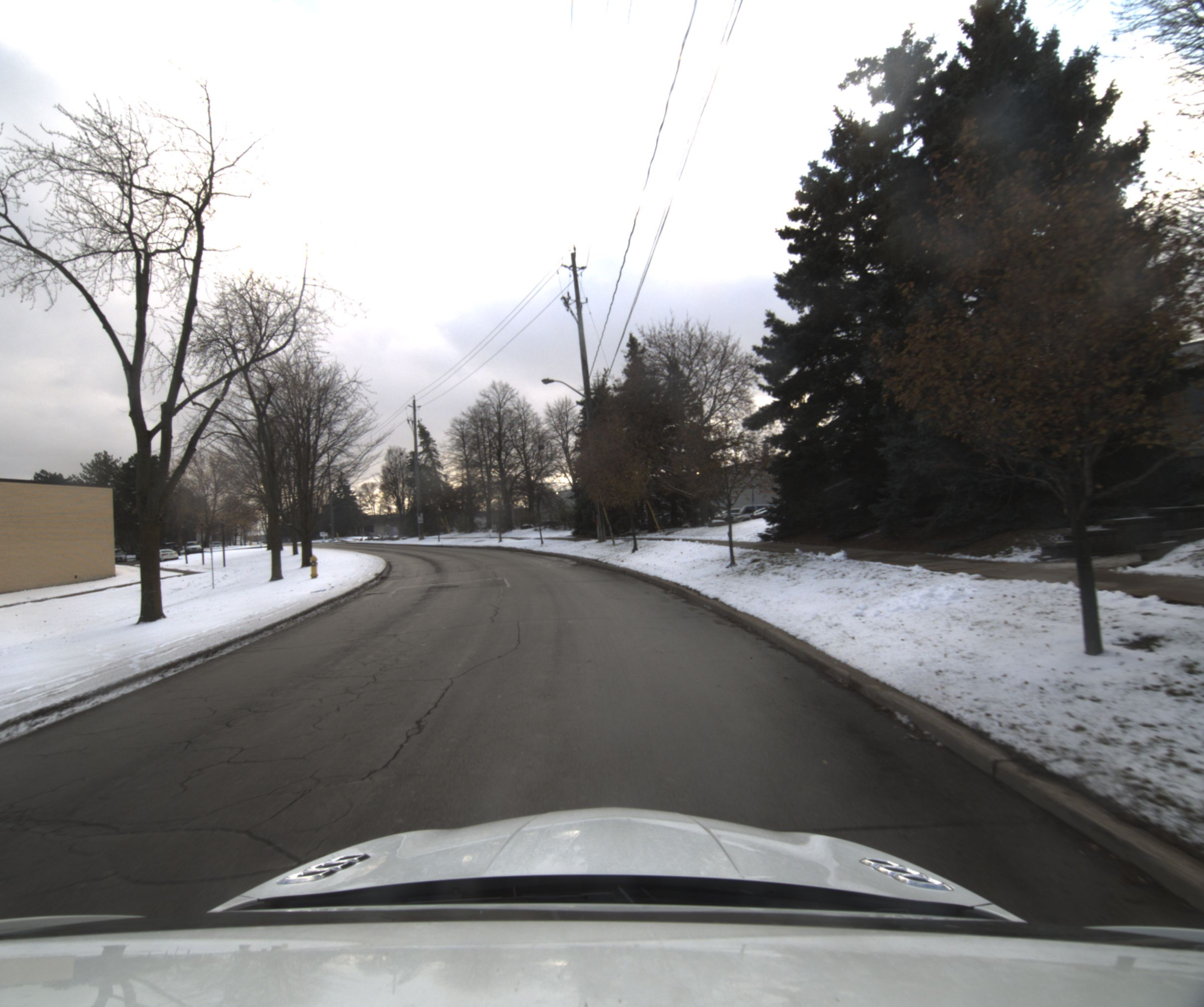}};
        \node[inner sep=0] (snow_img) at (0.5\linewidth,0)
            {\includegraphics[width=0.5\linewidth]{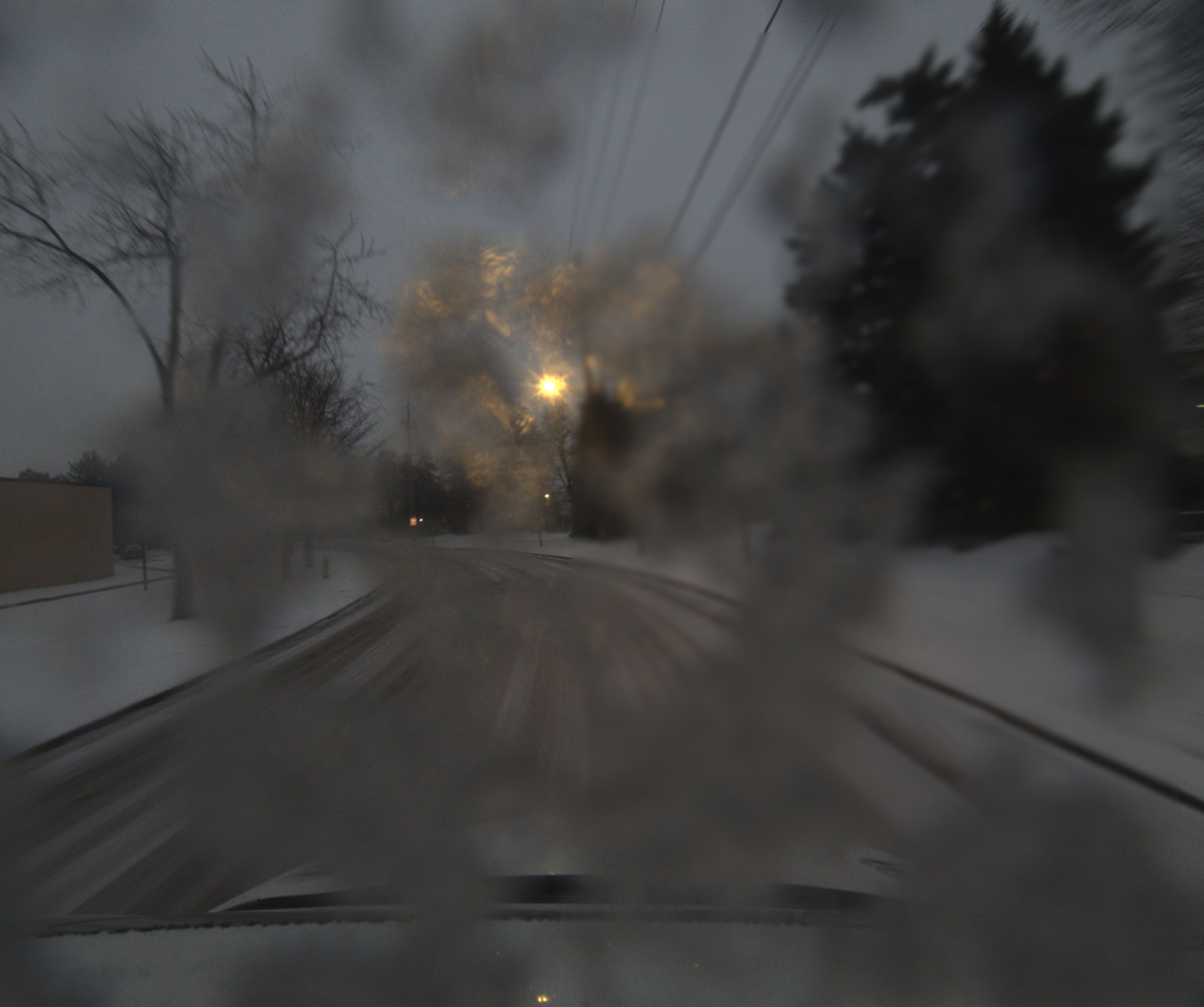}};

        \node[anchor=north west,
              fill=white, fill opacity=0.8,
              text opacity=1,
              inner sep=2pt] at (clear_img.north west)
              {\small mapping};
        \node[anchor=north west,
              fill=white, fill opacity=0.8,
              text opacity=1,
              inner sep=2pt] at (snow_img.north west)
              {\small localization};
        
        \coordinate (zSW) at
          ($(map.south west)!0.55!(map.south east)!0.38!(map.north west)$);
        
        \def\zoomSize{1.6cm}
        
        \coordinate (zNE) at ($(zSW) + (\zoomSize,\zoomSize)$);
        
        \draw[dashed, thick] (zSW) rectangle (zNE);

        \node[draw, dashed, thick, inner sep=1pt,
              anchor=north east] (zoomBig)
              at ($(map.north east)+(-0.6cm,-3.8cm)$)
        {
          \begin{tikzpicture}
            \clip (0,0) rectangle (3cm,3cm);
            \node[anchor=south west, inner sep=0pt] at (0,0)
              {\includegraphics[
                height=3cm,
                clip,
                trim=25mm 0mm 0mm 2mm
              ]{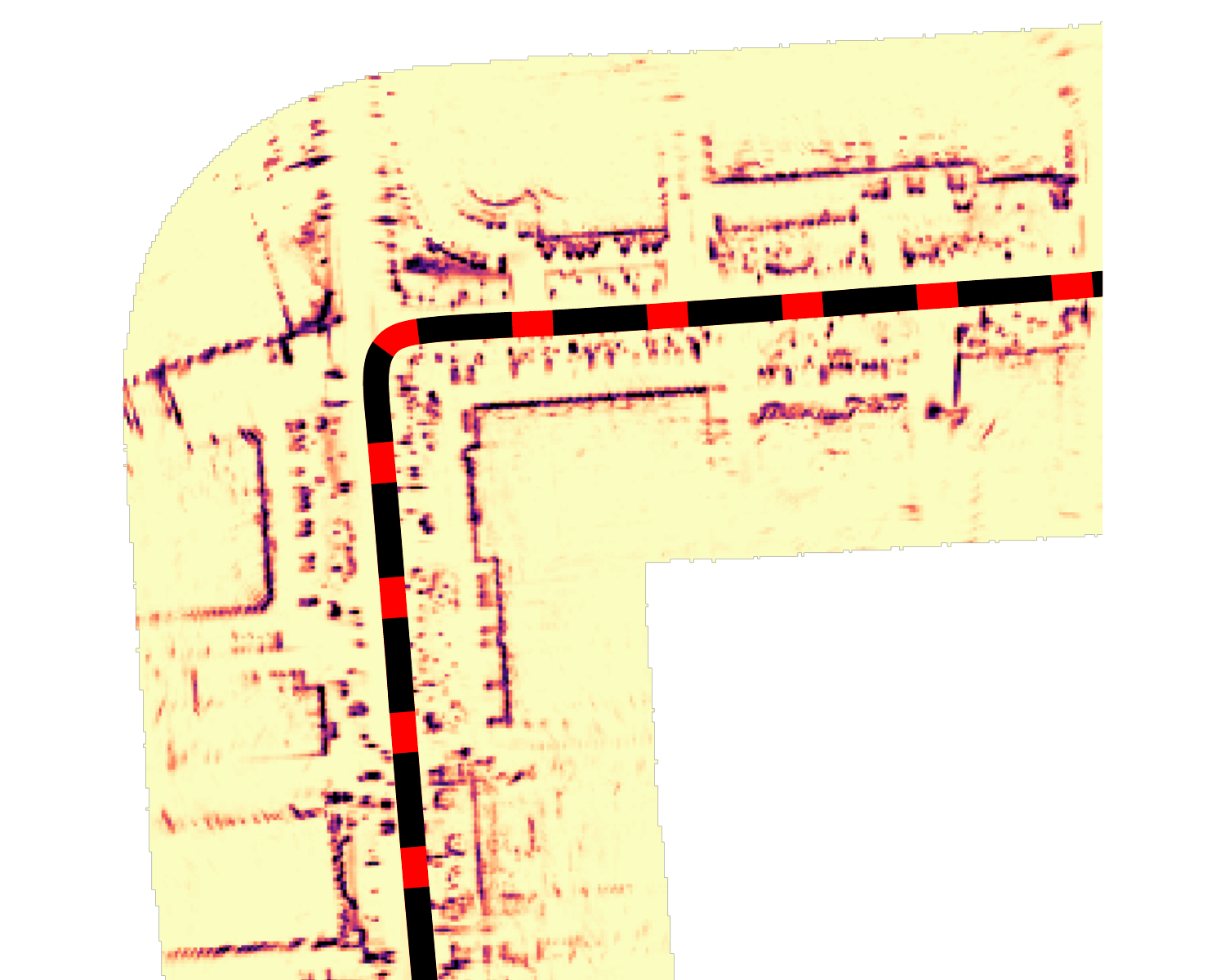}};
          \end{tikzpicture}
        };

        \draw[dashed, thick] (zNE) -- (zoomBig.north east);
        \draw[dashed, thick] (zSW) -- (zoomBig.south west);
        
    \end{tikzpicture}
    \caption{Dr-BA is a method for direct bundle adjustment of 2D radar data for simultaneous localization and mapping. Unlike pose-graph-based approaches, Dr-BA builds locally \& globally consistent maps directly for downstream tasks such as localization. The figure shows a Dr-BA-constructed map from a clear day with a localized (red) and ground truth (black) trajectory from a snowy day.}
    \label{fig:teaser}
    \vspace{-0.2cm}
\end{figure}

Dr-BA is a direct radar method.
Direct methods, as opposed to feature-based ones that extract sparse features from measurements, use all of the information provided by a sensor.
This is particularly valuable for map construction, since feature-based methods tend to produce relatively sparse maps that typically require downstream applications to use the same feature extraction as used for map construction.
Spinning radars lend themselves well to direct methods as they output dense range-azimuth-intensity polar images.
Previous radar mapping approaches (e.g., \cite{are_we_ready_for}) extracted point clouds from these images and grouped them using alignment techniques into point maps of the environment.
However, these point maps are generally very sparse, difficult to interpret, and require continued use of point cloud methods for localization.
In contrast, Dr-BA produces dense intensity maps that can be sampled at any frequency or location, facilitating downstream applications that may prefer maps of different densities.
Figure~\ref{fig:teaser} presents an example map produced by Dr-BA on a clear day, with a trajectory localized to that map using data collected during a snowstorm, demonstrating both the practicality of Dr-BA and radar’s robustness to adverse weather conditions.

One drawback of direct methods is that they can quickly become intractable as the number of measurements grows.
This is particularly problematic for \ac{ba} in autonomous driving, where maps span kilometres and are built from thousands of poses.
Leveraging the objective function from \cite{DSR} for medial imaging \ac{ba}, we formulate our problem in a way that allows us to separately estimate the sensor trajectory and the map.
In contrast to \cite{DSR}, we cast the problem as a general separable optimization and employ \ac{varpro} to decouple the objectives, avoiding the need for Jacobian-level analysis.
Consequently, Dr-BA complexity scales linearly with map size.

The key contributions of this paper are as follows:
\begin{itemize}
    \item Dr-BA: the first framework for direct radar \ac{ba}.
    \item The derivation of this framework as a separable optimization problem, separating the trajectory and map optimizations for efficiency.
    \item The demonstration of a downstream application with the implementation of a first \ac{drl} method using Dr-BA-generated maps.
    \item A publicly available ROS2-enabled implementation\footnote{Code available at \url{https://github.com/utiasASRL/dr_ba}.} of Dr-BA and DRL, validated on over $200\,\si{\km}$ of on-road data collected across diverse routes, including highly challenging environments.
\end{itemize}

\section{Related work}
This section provides a brief overview of both feature-based and direct methods for state estimation with 2D spinning (imaging) radars.
We refer the reader to \cite{nader2024survey, harlow2024newwave, venon2022millimeter} for a wider overview of radar techniques in robotics.

The term \emph{direct method} for state estimation originates from vision literature, where it is used to distinguish approaches that operate on the full set of sensor returns from feature-based methods that function on a subset of measurements.
This distinction also applies to 2D spinning radar-based methods.
Spinning radars typically return raw data in polar images with each row corresponding to an azimuth, each column to a range bin, and each pixel value to the recorded intensity.
The intensity represents the strength of radio-wave reflections from objects in the environment.
These images can be transformed into Cartesian coordinates for easier interpretability.

\subsection{Radar odometry methods}
The majority of spinning radar odometry methods to-date have been feature-based.
The authors of~\cite{cen2018precise} present a radar feature extraction and association method tailored for odometry.
ORORA~\cite{lim2023orora} introduces a graph-based pruning algorithm to remove outlier associations between features from two scans.
Other frameworks use \ac{ndt} to perform odometry~\cite{Kung2021AND}.
Modern radar-based methods often extract point clouds from the raw data before performing some \ac{icp}-like scan-to-scan or scan-to-map registration.
An extensive analysis of different point cloud extraction strategies for radar-based odometry is provided in \cite{preston2025finer}.
Some of the best-performing point-based radar odometry frameworks are~\cite{adolfsson2023cfear} that leverage point-to-distribution distances, and~\cite{burnett2025continuous} with point-to-point registration.
In addition to differing in the distance metrics used for registration, these methods rely on distinct state representations: discrete states in~\cite{adolfsson2023cfear} and continuous-time states in~\cite{burnett2025continuous}.
The choice of state representation leads to two strategies for handling motion and Doppler distortion in the radar data:~\cite{adolfsson2023cfear} performs a `one-off' undistortion step based on previous state estimates, while~\cite{burnett2025continuous} tightly couples the problem of data undistortion and ego-motion estimation in a single formulation.

While efficient, feature-based methods overlook a non-negligible amount of information from the original data.
In contrast, direct methods leverage the full radar return in Cartesian or polar form.
A common direct approach is to compute cross-correlation scores between consecutive radar scans using a discrete set of motion hypotheses and select the one with the highest score~\cite{masking_by_moving, Checchin2009, park2020pharao}.
It is important to note that many direct methods do not address the issues of motion and Doppler distortion of the raw data.
Using a radar unit with a particular frequency modulation pattern, it is possible to estimate the ego-velocity of the system with direct cross-correlation between consecutive collection azimuths~\cite{lisus2025doppler}.
This method provides a very efficient source of odometry when coupled with a gyroscope.
DRO~\cite{legentil2025dro} introduces a continuous cross-correlation optimization between a scan and a local map to directly address motion and Doppler distortion as part of odometry.

\subsection{Radar SLAM and localization methods}
While odometry is a required component in an autonomous system's stack, it is not sufficient for most applications.
For many tasks, and especially in self-driving, a critical feature is localization within previously built maps.
Localization along repeated routes does not require globally consistent maps, as demonstrated in~\cite{are_we_ready_for, furgale2010visual}.
However, the availability of global maps can enable a wider range of localization scenarios with more diverse trajectories.
In these scenarios, odometry alone cannot be used to generate maps of the environment due to its inherent drift.
\ac{slam} and place recognition methods are required to generate globally consistent maps.
Early work in \ac{slam} used radar for landmark-based state estimation~\cite{dissanayake2001slam} by extracting distinctive points and performing data association.
A similar feature-based approach, leveraging SIFT~\cite{lowe2004sift} from the vision-based research community, can be found in~\cite{callmer2011radarslamusingvisualfeatures}.
Pose-graph optimization (PGO) is the base of most existing radar \ac{slam} frameworks~\cite{adolfsson2023tbv, holder2019realtime, hong2022radarslam}.
These rely on loop-closure-detection approaches adapted from lidar applications \cite{he2016m2dp, himstedt2014largescale, kim2022scancontextpp} or specifically designed for radar place-recognition \cite{jang2023raplace, jang2025xpress}.
The most relevant PGO approach to this paper is Dr-PoGO \cite{legentil2026drpogo}, which employs a direct radar-based odometry method (DRO~\cite{legentil2025dro}), RaPlace~\cite{jang2023raplace} for loop-closure detection, and a modified version of DRO for loop-closure registration.
While PGO-based methods can provide fairly accurate trajectory estimates, the resulting maps, created from `stitching together' radar data at trajectory estimates, are not guaranteed to be locally consistent.
To-date, every \ac{slam} method except Dr-PoGO has been feature-base.

Dr-BA is the first direct radar \ac{ba} framework to produce maps that are both locally and globally consistent, optimizing direct registration across all collected scans.
We also introduce \ac{drl}, the first framework to leverage these maps for localization.

\subsection{Separable optimization}
In this paper, we leverage the theory of separable optimization to decouple the \ac{ba} problem.
Separable optimization problems admit an efficient closed-form solution for a subset of variables when the remaining variables are fixed.
One particularly effective approach for solving such problems is \ac{varpro}, which analytically eliminates the non-fixed variables and solves a reduced problem over the remaining ones \cite{golub_pereyra_separable, original_variable_projection}.
Recently, \ac{varpro} has been applied to several robotics algorithms, yielding improvements in both computational efficiency and convergence \cite{connor_varpro, papalia2025sparsevariableprojectionrobotic}.

Although the formulation in \cite{DSR} is not explicitly posed as a separable problem, the authors showed that direct \ac{ba} can be solved independently for the states and the map.
In this paper, we extend and generalize their result by explicitly casting the problem as a separable optimization.
This allows us to decouple the variables directly at the objective-function level, avoiding the need to inspect the Jacobian structure or rely on the Schur complement as done in \cite{DSR}.
Additionally, our formulation supports weighted error terms and simplifies the incorporation of additional error types.
We apply direct \ac{ba} to spinning radar data for the first time and demonstrate a simple, yet effective, localization framework that validates the resulting maps on a real downstream task.

\section{Methodology}
\label{sec:methodology}

The key idea of Dr-BA is to formulate the \ac{ba} error such that the map variables can be eliminated from the objective function.
This is done by considering the map variables to be intensities at predetermined sample locations, instead of the locations themselves as is common in feature-based \ac{slam}.
This formulation turns the problem into a separable optimization in which the poses and map intensities can be estimated independently through the use of \ac{varpro}, making trajectory estimation linear in the number of map states.
Furthermore, the entire map estimation problem can be simplified to a per-map-location estimation task, which is also linear in the number of map states.
A final practical benefit is that we are able to easily generate maps in a principled manner given poses from any source, be they the solution of the \ac{ba} problem or acquired by other means.

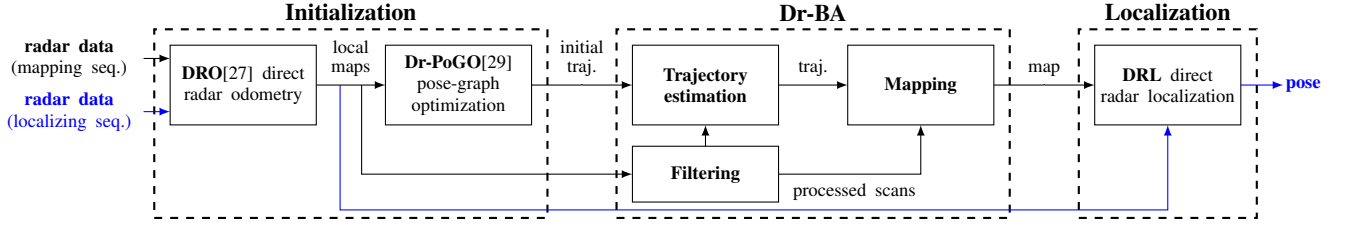
\begin{figure*}[ht!]
    \centering
    \def\hdist{2.6em}
    \def\hdistshort{1.0em}
    \def\vdist{0.8em}
    \def\blockheight{3.0em}
    \def\blockwidth{5.5em}
    \def\innerpad{0.6em}
    \def\midblockfactor{0.7}
    \def\textsize{\scriptsize}
    \begin{tikzpicture}[auto]
    \tikzstyle{input} = [draw=none, rectangle, minimum height = 2.5em, text width = 5.6em,  minimum width = 3.8em, inner sep=0, outer sep=0, align = center, node distance = 5em, execute at begin node=\setlength{\baselineskip}{8pt}]
    \tikzstyle{block} = [draw, fill=white, rectangle, minimum height = \blockheight, text width = \blockwidth,  minimum width = \blockwidth, align = center, inner sep=0, outer sep=0, node distance = 11em, execute at begin node=\setlength{\baselineskip}{8pt}] 
    \tikzstyle{wideblock} = [block, text width=((2*\blockwidth)+\hdist), minimum width=((2*\blockwidth)+\hdist)]
    \tikzstyle{midblock} = [block, text width=(\blockwidth), minimum width=(\blockwidth), minimum height = \midblockfactor*\blockheight]
    \tikzstyle{output} = [draw=none, fill=white, text=NavyBlue, rectangle, minimum height = 3em, text width = \blockwidth,  minimum width = \blockwidth, align = center, node distance = 11em, execute at begin node=\setlength{\baselineskip}{8pt}] 
    \tikzstyle{bigbox} = [draw, dashed, thick, fill=white, rectangle, text width = 0.3\columnwidth,  minimum width = 0.3\columnwidth, minimum height = (((1+\midblockfactor)*\blockheight)+(2*\innerpad)+\vdist), align = center, node distance = 11em, inner sep=0, outer sep=0, execute at begin node=\setlength{\baselineskip}{8pt}]

    \tikzstyle{varrowleft} = [text width = 5em, align=right, left, text width=((0.5*\blockwidth)+\innerpad), execute at begin node=\setlength{\baselineskip}{7pt}] 
    \tikzstyle{varrowright} = [text width = 5em, align=left, right, text width=((0.5*\blockwidth)+\innerpad), execute at begin node=\setlength{\baselineskip}{7pt}] 
    \tikzstyle{harrowabove} = [text width = \hdist, align=center, above, execute at begin node=\setlength{\baselineskip}{7pt}] 
    \tikzstyle{harrowbelow} = [text width = \hdist, align=center, below, execute at begin node=\setlength{\baselineskip}{7pt}] 

    \node [bigbox, text width = ((2*\blockwidth)+\hdist+(2*\innerpad)),  minimum width = ((2*\blockwidth)+\hdist+(2*\innerpad))] (init) {};
    \node [block, below right= 1.41*\innerpad of init.north west] (dro) {\textsize \textbf{DRO}\cite{legentil2025dro} direct radar odometry};
    \node [block, below left=1.41*\innerpad of init.north east] (pogo) {\textsize \textbf{Dr-PoGO}\cite{legentil2026drpogo} pose-graph optimization};

    \node [input, left=\hdistshort of dro.160] (radarmap) {\textsize \textbf{radar data}\\(mapping seq.)};
    \node [input, left=\hdistshort of dro.200, text=blue] (radarloc) {\textsize \textbf{radar data}\\(localizing seq.)};

    \node [bigbox, right=\hdist of init, text width = ((2*\blockwidth)+\hdist+(2*\innerpad)),  minimum width = ((2*\blockwidth)+\hdist+(2*\innerpad))] (ba) {};
    \node[block, below right=1.41*\innerpad of ba.north west] (traj) {\textsize \textbf{Trajectory estimation}};
    \node[block, below left=1.41*\innerpad of ba.north east] (mapping) {\textsize \textbf{Mapping}};
    \node[midblock, above right=1.41*\innerpad of ba.south west] (filter) {\textsize \textbf{Filtering}};

    \node [bigbox, text width = ((\blockwidth)(2*\innerpad)),  minimum width = ((\blockwidth)+(2*\innerpad)), right=\hdist of ba] (loc) {};
    \node [block, below right= 1.41*\innerpad of loc.north west] (drl) {\textsize \textbf{DRL} direct radar localization};

    \node [input, right=\innerpad+\hdistshort of drl, text=blue, minimum width = 0em, text width=1.5em] (pose) {\textsize \textbf{pose}};

    \draw[->] (radarmap) -- (radarmap.east -| dro.west);
    \draw[->, blue] (radarloc) -- (radarloc.east -| dro.west);
    \draw[->] (dro) -- node[harrowabove]{\textsize local maps} (pogo);
    \draw[->] ($(dro)!0.55!(pogo)$) |- (filter);
    \draw[->,blue] ($(dro)!0.45!(pogo)$) |-  ([yshift=-(((1+\midblockfactor)*\blockheight)-\blockheight+(0.5*\innerpad)+\vdist)]drl.south) -- (drl.south);

    \draw[->] (pogo) -| ($(pogo)!0.5!(traj)$) node[harrowabove]{\textsize initial traj.} |-(traj);
    \draw[->] (traj) -- node[harrowabove]{\textsize traj.} (mapping);
    \draw[->] (filter) -- (traj);
    \draw[->] (filter) -| node[harrowbelow, text width=6em, xshift=-2.5em]{\textsize processed scans} (mapping);

    \draw[->] (mapping) -| ($(drl)!0.5!(mapping)$) node[harrowabove]{\textsize map} |- (drl);
    \draw[->, blue] (drl) -- (pose);

    \node[above=0.3em of init, inner sep=0em] {\small \textbf{Initialization}};
    \node[above=0.3em of ba, inner sep=0em] {\small \textbf{Dr-BA}};
    \node[above=0.3em of loc, inner sep=0em] {\small \textbf{Localization}};
    
\end{tikzpicture}
    \caption{An overview of the implementation of Dr-BA, including initialization steps and a downstream localization task.}
    \label{fig:method}
\end{figure*}

\subsection{Direct bundle adjustment}
We wish to simultaneously estimate the intensity values $\mbf{i} \in \Real^P$ of a set of $d$-dimensional map points, $\map~=~\{\mbf{m}_v~\in~\Real^d\}_{v=1}^{P}$, and the trajectory of a sensor within that map.
The intensity value at each map position, $\mbf{m}_v$, is given by $i(\mbf{m}_v)~\equiv~i_v$.
Note that we are not optimizing for the positions $\mbf{m}_v$ themselves, only for the intensities at those positions.
We intentionally keep the map sampling general, since it is application- and compute-resource-dependent.
In this paper, we use a uniform sampling approach, but one could employ more complex information-density-based sampling depending on the task at hand.
The trajectory is represented as a set of $d$-dimensional discrete states, $\mbf{T} = \{\mbf{T}_n\}_{n=1}^N$, at which intensity measurements, called \textit{scans}, are rooted.
We can get the measured intensity value from a given scan at $\mbf{m}_v$ by using a measurement sampling function $\beta(\cdot)$,
\begin{align}
    \tilde{i}_{v, n} &= \beta(\mbf{m}_v, \mbf{T}_n).
\end{align}
Depending on the scan representation, $\beta(\cdot)$ can directly query or interpolate a measurement at $\mbf{m}_v$ in the scan.

We form a general loss on intensity differences across the map as a weighted least-squares problem
\begin{align}
    \mbf{i}^\star, \mbf{T}^\star &= \argmin_{\mbf{i}, \mbf{T}} \sum_{v=1}^{P} \sum_{n\in\vizset_{v}}^{} w_{v, n} \left(i_v - \tilde{i}_{v, n} \right)^2,\label{eq:full_ba_objective}
\end{align}
where $\vizset_v$ is the set of indices of discrete states that observe a given $\mbf{m}_v$, and $w_{v, n} > 0$ is the weight of each residual.

We direct the reader's attention to the fact that the intensities, $i_v$, enter linearly into the error terms of the cost. This fact allows us to apply the \ac{varpro} method to this problem, which can eliminate such linear variables to reduce problem complexity \cite{original_variable_projection}.
Treating $\mbf{T}$ as fixed and optimizing only over $\mbf{i}$ in \eqref{eq:full_ba_objective}, we note that the optimal value of each intensity, $i_v$, is given by
\begin{align}
    i_v(\mbf{T}) &= \frac{1}{w_v}\sum_{n\in\vizset_{v}}^{} w_{v, n}\tilde{i}_{v, n} \equiv \bar{i}_{v},\label{eq:i_as_func_pose}
\end{align}
where $\sum_{n\in\vizset_{v}} w_{v, n} \equiv w_v$ for brevity.
Thus, if the scan poses are known, the intensity values of the map will simply be the weighted average of the interpolated scan intensities that overlap each map location.
Conveniently, \eqref{eq:i_as_func_pose} enables efficient, easily parallelized per-location map intensity generation, avoiding the need to form a large linear system.

Having solved for $\mbf{i}$ as a function of $\mbf{T}$, we plug \eqref{eq:i_as_func_pose} back into \eqref{eq:full_ba_objective} to now optimize only over $\mbf{T}$ as
\begin{align}
    \mbf{T}^\star &= \argmin_{\mbf{T}} \sum_{v=1}^{P} \sum_{n\in\vizset_{v}}^{} w_{v, n} \left(\bar{i}_{v} - \tilde{i}_{v, n} \right)^2, \label{eq:pose_optimization}
\end{align}
where the objective function is still dependent on sampling locations, $\mbf{m}$, but where our substitution has projected out the explicit dependence on the intensities, $\mbf{i}$.
Note that \eqref{eq:pose_optimization} is finding an optimal $\mbf{T}$ such that the weighted variance of intensity measurements is minimized at all map locations.
Since \eqref{eq:pose_optimization} is a fully relative problem, we fix the first pose.

We iteratively solve \eqref{eq:pose_optimization} using Gauss-Newton, alternating between linearization of the objective and solving the resulting linear system as
\begin{equation}
    \mbf{H}\,\Delta\mbf{T} = \mbf{b},
    \label{eq:linear_problem}
\end{equation}
where $\mbf{H} = \mbf{J}^\top\mbf{J} \in \Real^{dN\times dN}$, $\mbf{b}=-\mbf{J}^\top\mbf{e} \in \Real^{dN}$, $\mbf{e}$ is the vector of residuals, $\mbf{J}$ is the Jacobian of the residuals with respect to the state perturbations, and $\Delta\mbf{T}$ is the pose update for a given iteration.
Both $\mbf{H}$ and $\mbf{b}$ can be efficiently computed as
\begin{align}
    \mbf{H} = \sum_{v=1}^{P} \mbf{J}_{v}^\transpose \mbf{J}_{v}, \label{eq:H_comp}\quad
    \mbf{b} = \sum_{v=1}^{P} \mbf{J}_{v}^\transpose \mbf{e}_{v},
\end{align}
where $\mbf{J}_{v} \in \Real^{\vert\vizset_{v}\vert\times dN}$ is the Jacobian of error terms involving only $\mbf{m}_v$, grouped as $\mbf{e}_v \in \Real^{\vert\vizset_{v}\vert}$, with respect to all states.
In the worst case, solving $\mbf{H}$ has cubic complexity in the number of trajectory poses.
However, in practice, $\mbf{J}_{v}$ is sparse in many bundle adjustment problems since only $d\,|\mathcal{V}_v|$ columns are non-zero.
In typical mapping scenarios, co-observability between states is much smaller than the total number of states, so $|\mathcal{V}_v| \ll N$.
Consequently, $\mbf{H}$ also becomes relatively sparse with entries only on the block diagonal and in cross-correlation blocks between states observing the same map locations.
Notably, the size of the linear problem~\eqref{eq:linear_problem} does not depend on the number of map points sampled.
Increasing the sampling rate of the map will affect the number of loops we execute, but not the amount of memory required to solve for the trajectory.
After solving for $\mbf{T}$, we can use \eqref{eq:i_as_func_pose} to build the map.

Note that we have yet to specify a map, scan, or state representation, nor have we defined a specific scan sampling function.
We apply this efficient \ac{ba} approach to the task of direct 2D spinning radar mapping in this paper, referring to our method as Dr-BA.
However, this formulation can be used with any sensor that produces intensity measurements in any number of dimensions.

The derivation extends naturally to objectives of the form \eqref{eq:full_ba_objective} with additional pose- or map-only terms, as these do not affect separability. For instance, state priors for full SLAM can be incorporated directly into \eqref{eq:pose_optimization} without affecting \eqref{eq:i_as_func_pose}.

\subsection{Direct localization}
The simplest way to do localization, given our previous formulation, is to re-use \eqref{eq:full_ba_objective} to optimize exclusively for a single state given fixed map intensity values $\mbf{i}$.
The objective function to localize pose $\mbf{T}_n$ in the map is thus
\begin{align}
    \mbf{T}_n^\star &= \argmin_{\mbf{T}_n} \sum_{v = 1}^{Q} w_{v, n} \left(i_v - \tilde{i}_{v, n}\right)^2,\label{eq:loc_objective}
\end{align}
where we loop over the $Q$ intensities observable by the scan at $\mbf{T}_n$.
Just as with the BA problem, \eqref{eq:loc_objective} is general across any map, scan, and state representation, as well as any scan sampling scheme.
Gauss-Newton can again be used to iteratively solve for $\mbf{T}_n$.
When used with radar data, we term this method \ac{drl}.

\begin{figure}
    \centering
    \begin{tikzpicture}[remember picture]

    \node (fig) {
    \scalebox{0.8}{
    \begin{minipage}{\linewidth}
        \centering
        \setlength{\fboxsep}{0pt} 
        \fbox{\includegraphics[decodearray={1 0}, width=0.5\linewidth]{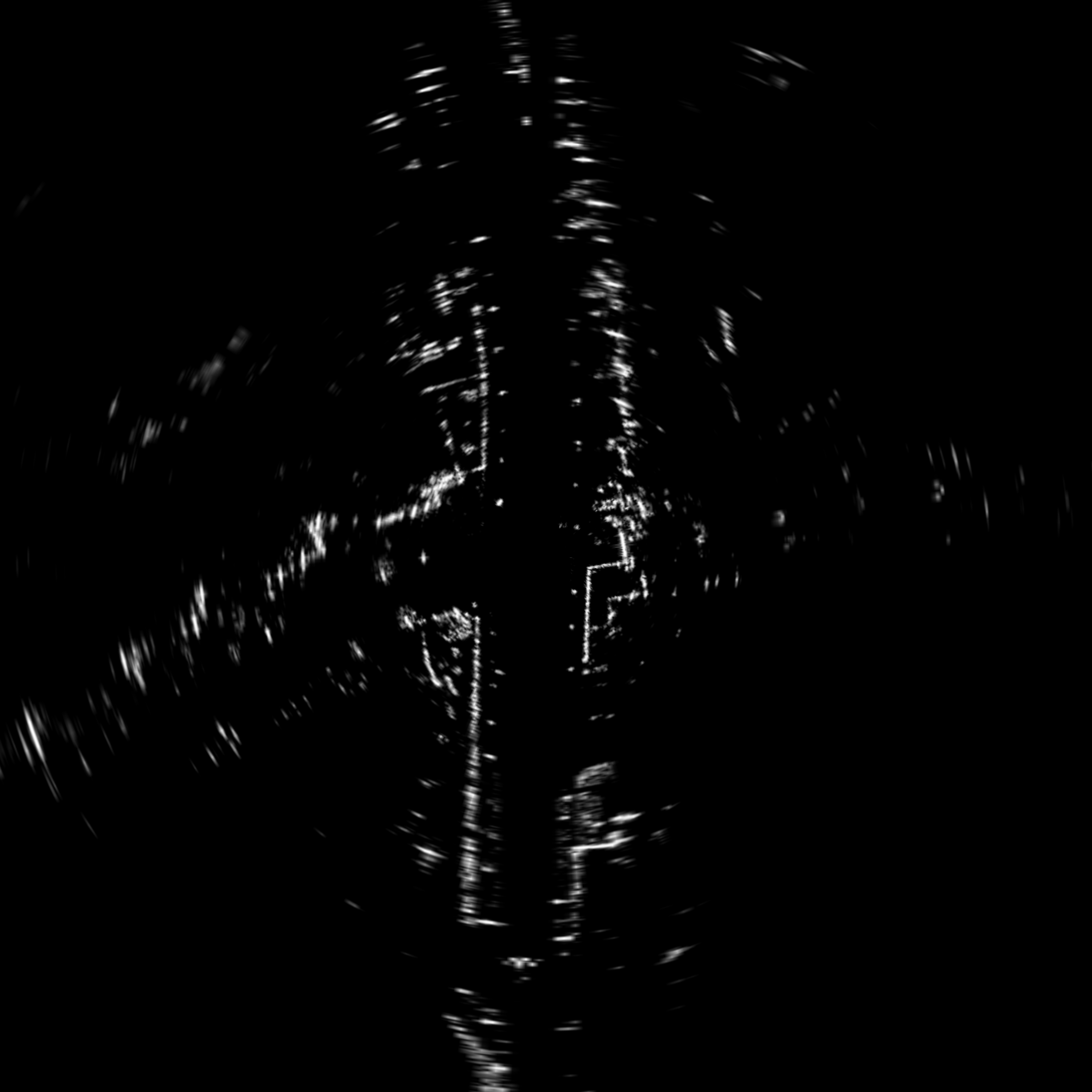}}%
        \fbox{\includegraphics[decodearray={1 0}, width=0.5\linewidth]{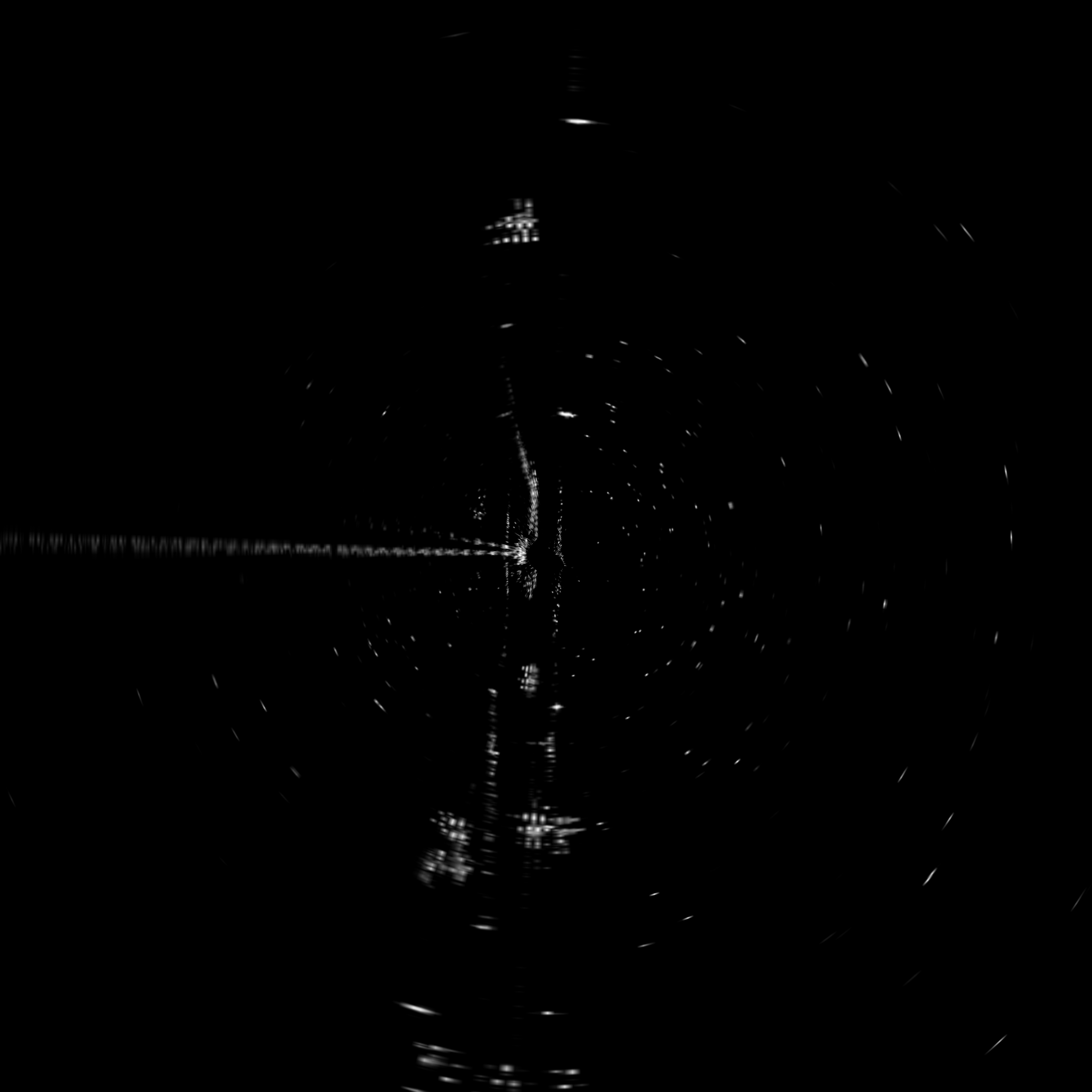}}
        \\[-0.3pt]
        \fbox{\includegraphics[decodearray={1 0}, width=0.5\linewidth]{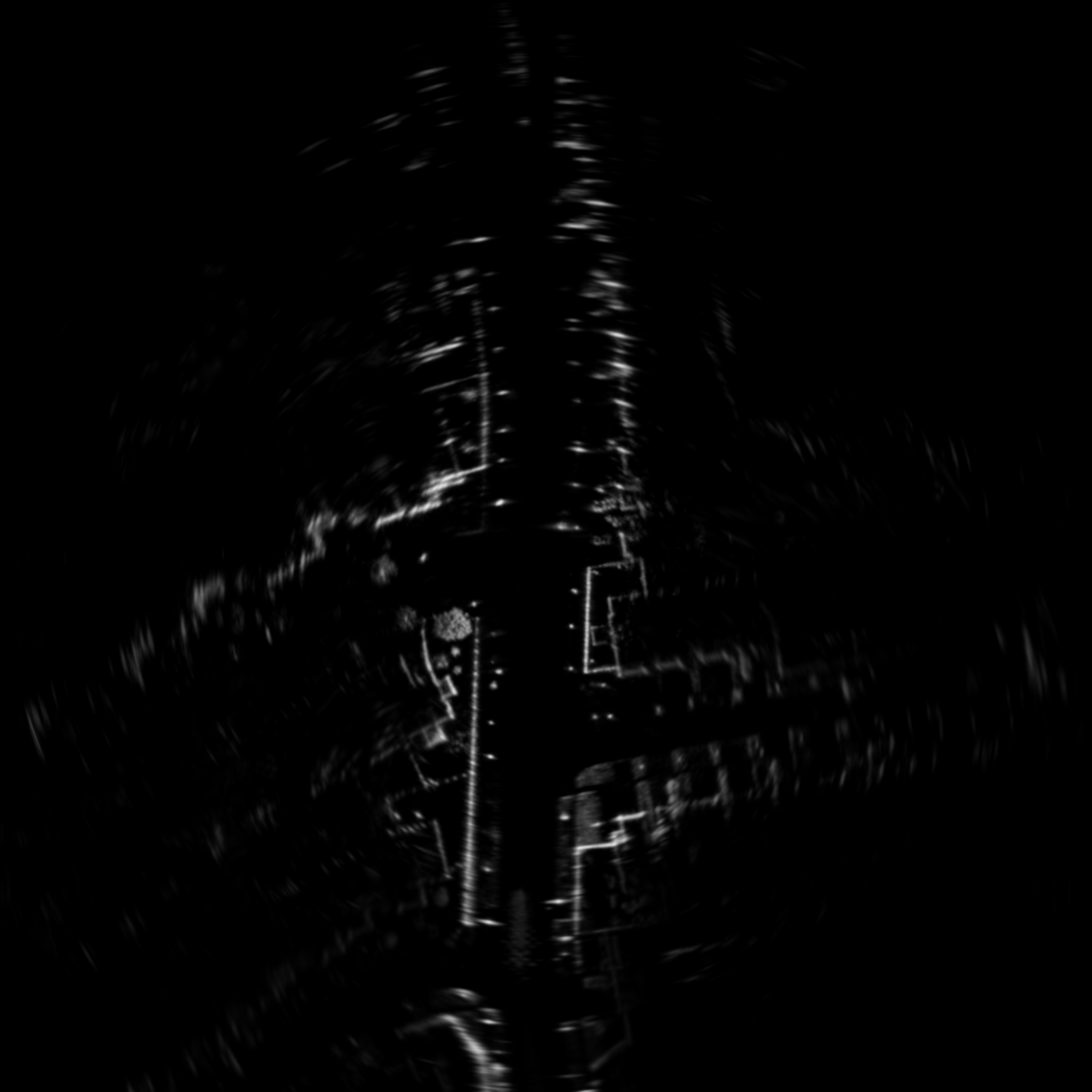}}%
        \fbox{\includegraphics[decodearray={1 0}, width=0.5\linewidth]{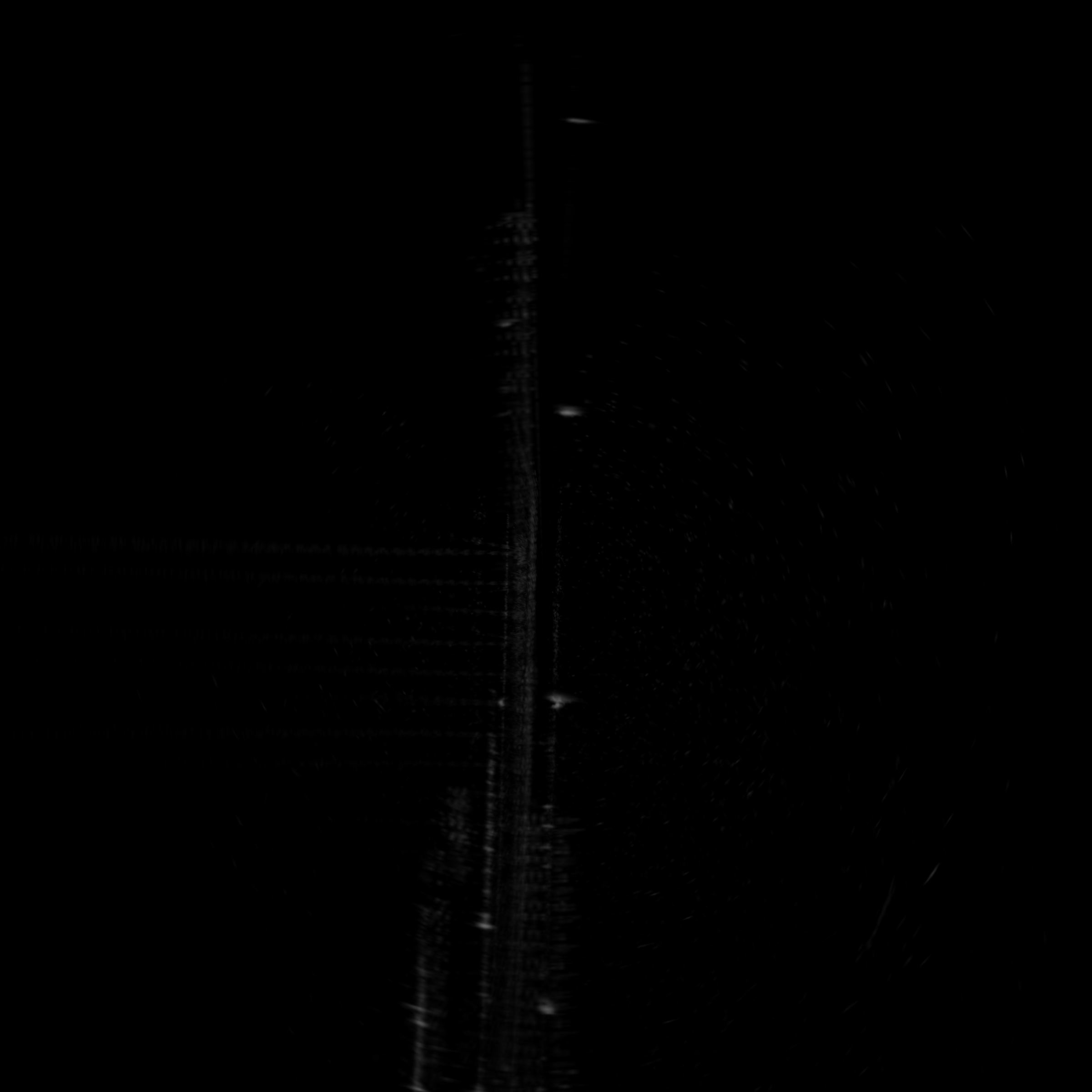}}
        \\[-0.3pt]
        \fbox{\includegraphics[decodearray={1 0}, width=0.5\linewidth]{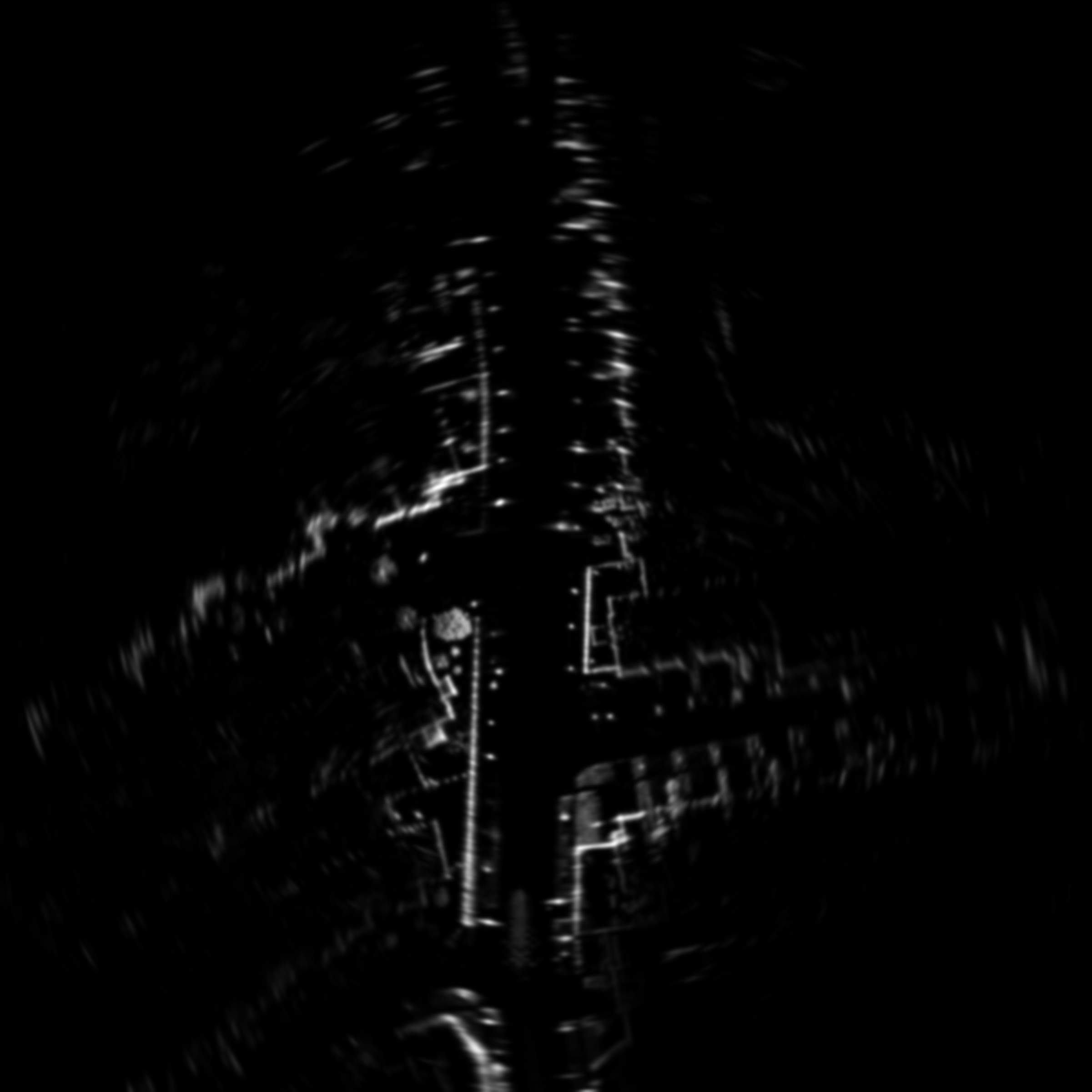}}%
        \fbox{\includegraphics[decodearray={1 0}, width=0.5\linewidth]{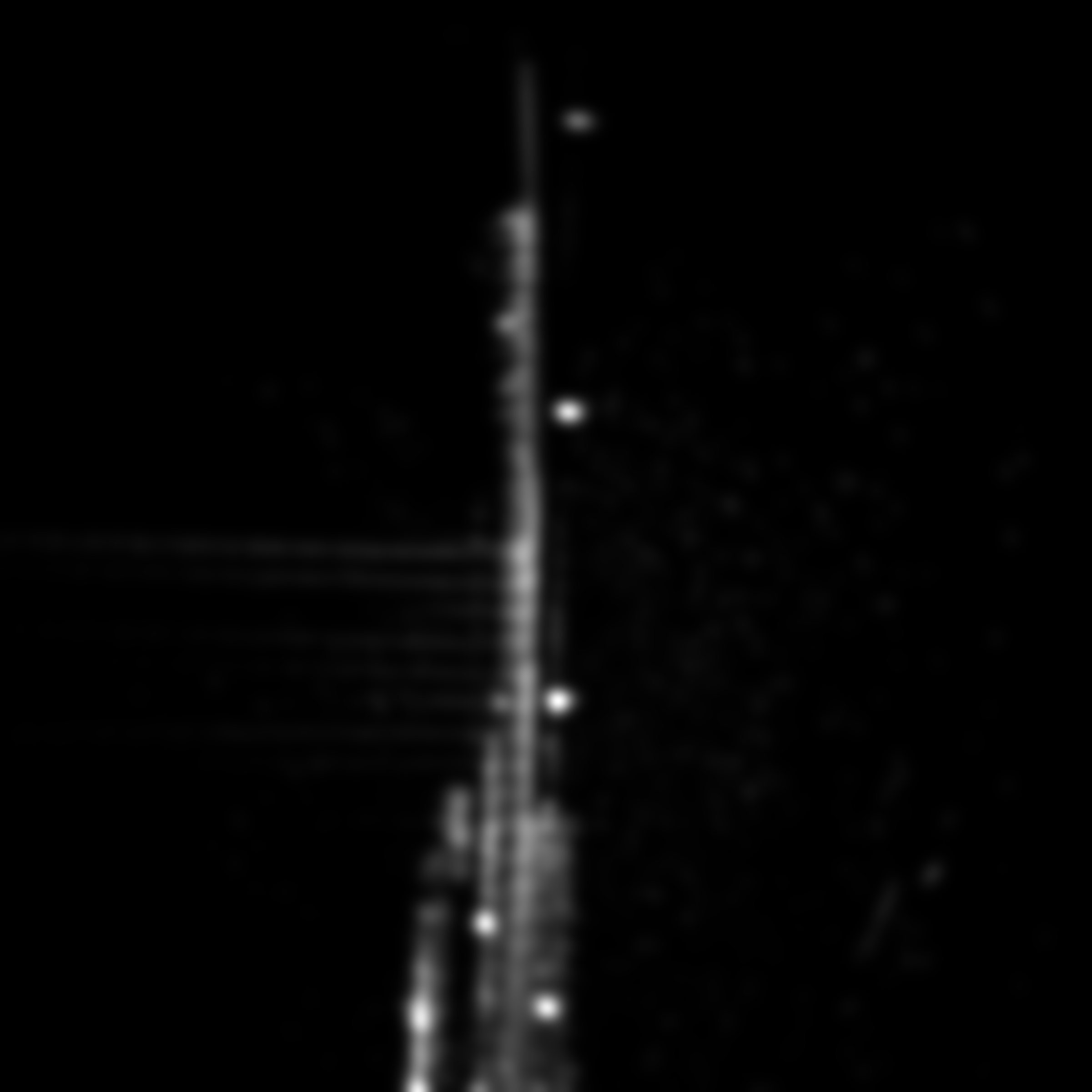}}
        \\[-0.3pt]
        \fbox{\includegraphics[decodearray={1 0}, width=0.5\linewidth]{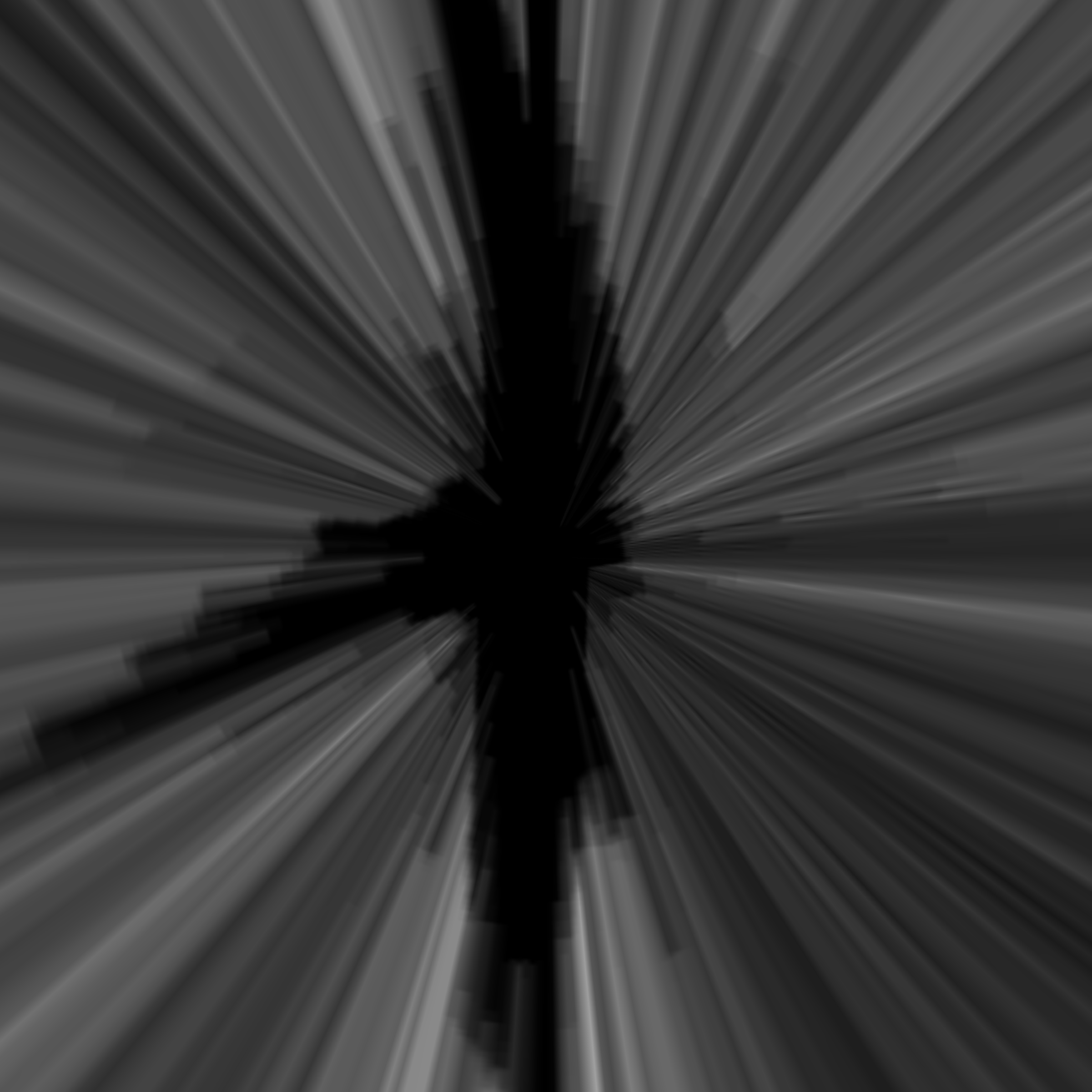}}%
        \fbox{\includegraphics[decodearray={1 0}, width=0.5\linewidth]{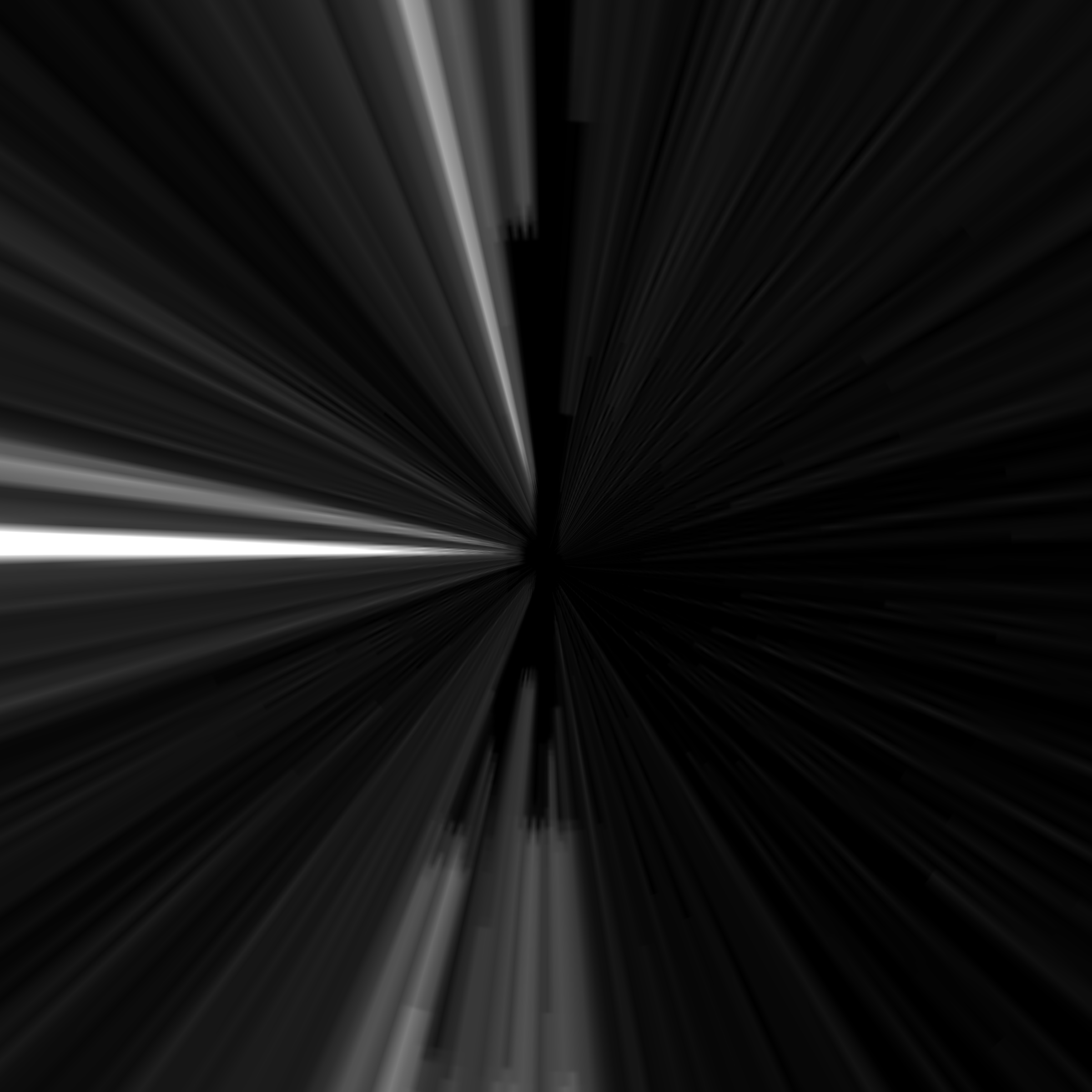}}
    \end{minipage}
    }
    };

    \node[rotate=90, anchor=center] 
      at ($(fig.north west)+( -0.1cm,-1.9cm)$) {Radar scans};
    
    \node[rotate=90, anchor=center] 
      at ($(fig.north west)+( -0.1cm,-5.4cm)$) {Local maps (DRO)};
    
    \node[rotate=90, anchor=center] 
      at ($(fig.north west)+( -0.1cm,-9.0cm)$) {Adaptive blurring};
    
    \node[rotate=90, anchor=center] 
      at ($(fig.north west)+( -0.1cm,-12.6cm)$) {Cumulative images};
    
    \end{tikzpicture}

    \caption{Different processing steps for leveraging raw radar images for Dr-BA. Data from an `informative' suburban location is shown on the left and data from an `uninformative' skyway location is shown on the right.}
    \label{fig:radar_data_preprocessing}
\end{figure}

\section{Implementation}
\label{sec:implementation}

The general theory from Section~\ref{sec:methodology} is applied explicitly in the formulation below; an implementation overview is shown in Figure~\ref{fig:method}.
We perform \ac{ba} directly on spinning radar images to construct dense, globally consistent maps for downstream autonomous driving tasks such as localization.
Specifically, we fuse many 2D radar images, such as those shown at the top of Figure~\ref{fig:radar_data_preprocessing}, collected along an on-road \textit{sequence} into a single map (Figure~\ref{fig:teaser}).

\subsection{Map Representation}
Since the radar collects 2D data, the map is 2D with locations $\mbf{m} \in \Real^2$ and intensities $i \in [0, 1]$.
For simplicity, we sample it regularly with resolution $r_v$.

\subsection{State Representation}
We consider the radar data to be collected at discrete poses, $\mbf{T} = \{\mbf{T}_n \in SE(2)\}_{n=1}^N$, each defining the transform from the radar to the world frame at that timestamp.
We optimize the poses, $\mbf{T}$, directly on-manifold and select the moving-frame-invariant perturbation scheme to compute Jacobians and iteratively optimize our state \cite{barfoot2024state}.

In order to not over-index on stationary segments during which many repeated scans are collected, a keyframing approach, common in visual \ac{ba} \cite{ptam}, is taken for selecting poses.
A new keyframe, and consequently scan, is only added to the problem if sufficient motion ($\geq 5 \, \si{\m}$ or $\geq30\si{\degree}$) has been experienced since the last keyframe.

As Dr-BA performs no explicit loop-closure or feature-based data association, a good initial trajectory estimate is required for convergence.
We initialize using Dr-PoGO \cite{legentil2026drpogo}.

\subsection{Scan Representation}
\label{sec:scan_representation}
We assume the scans to be Cartesian images with a resolution $r_r$ per $(u,v)$ pixel.
Following the standard pixel coordinate convention with the origin in the top-left corner, the positive $u$ axis toward the right, and the positive $v$ axis toward the bottom, a map point $\mbf{m}_v$ in the world frame is converted to pixel coordinates $\mbf{p}_v^n$ of a radar scan captured from $\mbf{T}_n$ as
\begin{align}
    \mbf{p}_v^n &= \begin{bmatrix}
        0 & \frac{1}{r_r} & 0 \\ -\frac{1}{r_r} & 0 & 0
    \end{bmatrix} \mbf{T}_n\inv\begin{bmatrix}\mbf{m}_v\\1\end{bmatrix} + \frac{1}{2}\begin{bmatrix}
        w - 1 \\ h - 1
    \end{bmatrix},
    \label{eq:map_to_pixel}
\end{align}
with $w$ and $h$ the width and height of the image.
As the pixel position from~\eqref{eq:map_to_pixel} is unlikely to be a whole number, we use bilinear interpolation to obtain $\tilde{i}_{v, n}(\mbf{p}_v^n)$.
The uncertainty of $\tilde{i}$, used to compute the corresponding $w_{v, n}$ in \eqref{eq:full_ba_objective} and \eqref{eq:loc_objective}, is formed from a constant per-pixel uncertainty ($\mathcal{N}(\mbf{0}, 0.1^2)$) and a range-dependent uncertainty ($\mathcal{N}(\mbf{0}, (0.005 \norm{\mbf{m}_n})^2)$, where $\mbf{m}_n$ is the position of $\mbf{m}_v$ in radar frame $\mbf{T}_n$).

The proposed \ac{ba} formulation optimizes the poses of the sensor for a discrete set of keyframes.
However, spinning radars do not take instantaneous snapshots of the environment; they sweep their surroundings with one or more dishes over a period typically between 100 and 500 $\si{\ms}$.
Accordingly, a radar scan collected during a $360\si{\degree}$ sweep is subject to motion distortion.
Additionally, when the sensor is moving relatively to its environment, measurements will also be subject to the Doppler effect.
To obtain undistorted radar scans, both in terms of motion and Doppler effect, we again leverage DRO~\cite{legentil2025dro}.
DRO outputs undistorted scans and smooth local maps as Cartesian images.
Local maps are used as inputs to Dr-BA due to their inherent noise rejection and the larger convergence basin induced by slight smoothing.

However, even with the smoothing offered by local maps, radar measurements collected in very sparse environments (see Figure~\ref{fig:radar_data_preprocessing}) can contain almost no gradient information for pixel-level optimization.
To address this, we adopt an adaptive Gaussian blurring approach.
For each scan, we compute the amount of `valuable information' within the scan by a simple check of the percent of pixels that contain intensities above a threshold (0.5).
If the percent is lower than some bound ($0.3\%$), we apply a progressively stronger Gaussian blur to the image until the bound is reached.
This dynamically changes the amount of blurring applied to scans throughout a sequence, keeping informative scans almost untouched and heavily blurring scans that need a wider convergence basin around the few available features.
Dynamic blurring is the key to getting good alignment in sparse environments such as the one visualized on the right of Figure~\ref{fig:map_comparison}.

Another practical problem with direct radar \ac{ba} is that objects in the environment can prevent the radar signal from measuring to its full range, despite the radar image always returning the same number of azimuth-range-intensity measurements.
This means that some zero-intensity measurements correspond to a lack of material in the environment, while others correspond to regions that the radar signal never reached due to occlusions.
In the latter case, incorporating these `fake' zero measurements yields an incorrect down-weighting of the corresponding map locations.
A related inverse problem is radar saturation, often caused by `ringing', where the signal repeatedly reflects between parallel surfaces.
This ringing, visually seen in the right radar scan in Figure~\ref{fig:radar_data_preprocessing}, returns high intensities all along a given azimuth regardless of what is physically there.
These returns incorrectly up-weight those regions.
To handle both of these cases, we construct a `cumulative' radar scan based on the input to Dr-BA.
For each azimuth, we sum the intensities outward along each ray and store the cumulative sum in the corresponding pixel of the cumulative image.
We then ignore all zero measurements that have a cumulative score above a threshold (0.2) and all measurements with a score above another higher threshold (0.9).

The dynamic smoothing and cumulative image processing are handled as part of the Filtering block in Figure~\ref{fig:method}.
Visual examples of the different types of scans that take part in the Dr-BA pipeline are shown in Figure~\ref{fig:radar_data_preprocessing}.
Finally, the impact on performance from these different parts of the pre-processing is studied in Section~\ref{sec:ablation}.

\section{Experiments}

\begin{table*}[ht]
    \centering
    \caption{Average ATE and EPE (all [\si{\m}]) per sequence type for TBV-SLAM, DRO, Dr-PoGO, and Dr-BA trajectories.}
    \setlength{\tabcolsep}{2pt}
    \small
    \begin{tabularx}{\linewidth}{lYYYYY}
        \toprule
        \textbf{Method} & \texttt{Suburbs} & \texttt{Industrial} & \texttt{Skyway} & \texttt{Forest} & \texttt{Farm}
        \\
        \midrule
        DRO \cite{legentil2025dro} & 9.53 / 38.9 & 6.32 / 17.2 & 18.6 / 60.7 & 25.8 / 80.5 & 38.8 / 170.4
        \\
        TBV-SLAM \cite{adolfsson2023tbv} & 6.11 / 0.55 & 4.72 / 0.82$^1$ & - / -$^4$ & - / -$^4$ & - / -$^4$
        \\
        Dr-PoGO \cite{legentil2026drpogo} & 0.75 / 0.42 & 1.58 / 0.77 & 3.16 / 0.23 & \textbf{4.31} / \textbf{0.75} & \textbf{4.19} / 0.56
        \\
        Dr-BA (Ours) & \textbf{0.54} / \textbf{0.25} & \textbf{1.31} / \textbf{0.31} & \textbf{2.46} / \textbf{0.17} & 4.36 / 0.78 & 4.24 / \textbf{0.42}
        \\
        \bottomrule
        \multicolumn{6}{l}{\scriptsize Results reported as XX / YY, with XX the ATE and YY the EPE.}
        \\
        \multicolumn{6}{l}{\scriptsize $^1$ The superscript indicates the number of failed sequences (ATE above 1\% of the sequence distance).}
    \end{tabularx}
    \label{tab:traj_eval}
\end{table*}

\subsection{Dataset}
To demonstrate the performance of Dr-BA and \ac{drl}, we use a subset of the Boreas Road Trip (Boreas-RT) dataset~\cite{lisus_brrt26}, which contains data from a Navtech RAS6 radar, a Silicon Sensing DMU41 6-DoF IMU, and an Applanix RTK-GNSS/INS solution for ground-truthing.
The radar produces scans with a resolution of $r_r~=~\SI{0.0438}{\m}$ and a range of up to $\SI{300}{\m}$, although only the first $\SI{100}{\m}$ have useful returns for \ac{ba} in practice.
The subset consists of 20 sequences in 5 different environments (4 repeats per route) for a total of $\SI{206.4}{\km}$: \texttt{Suburbs} (\SI{7.9}{\kilo\meter}), \texttt{Industrial} (\SI{5.4}{\kilo\meter}), \texttt{Skyway} (\SI{11.1}{\kilo\meter}), \texttt{Forest} (\SI{16.4}{\kilo\meter}), and \texttt{Farm} (\SI{10.8}{\kilo\meter}).
The \texttt{Suburbs} route goes along suburban streets before returning to the starting location along the same set of streets in the opposite direction.
The \texttt{Industrial} route loops through an industrial area with commercial buildings before coming back to the starting location, with some sequences collected during a snowstorm.
\texttt{Skyway} sequences were collected driving up and down a busy high-speed skyway with very limited structural features but a lot of measurement overlap, making both the problem of odometry and loop-closure detection/registration extremely challenging.
The \texttt{Forest} and \texttt{Farm} routes form large loops in very challenging environments with very few buildings along the vehicle's path and not a lot of measurement overlap.
The \texttt{Suburbs} and \texttt{Industrial} sequences represent `easy', structured on-road conditions that are typically tested by autonomous vehicle \ac{slam} methods.
The \texttt{Skyway}, \texttt{Forest}, and \texttt{Farm} sequences represent environments in which current \ac{sota} algorithms are expected to perform poorly, and in some cases fail.

\begin{figure}
    \centering
    \def\imgwidth{0.80} 
    \def\legendspace{0mm}
    \def\rowspace{5mm}
    \def\trimsize{11}
    \def\trimsizeB{9}
    \begin{tikzpicture}[remember picture]
    \node (fig) {
    \scalebox{0.95}{
      \begin{minipage}{\linewidth}
        \centering
    
    \begin{minipage}[t]{0.48\linewidth}
        \centering
        \includegraphics[clip, trim=\trimsize cm \trimsize cm \trimsize cm \trimsize cm,width=\imgwidth\linewidth]{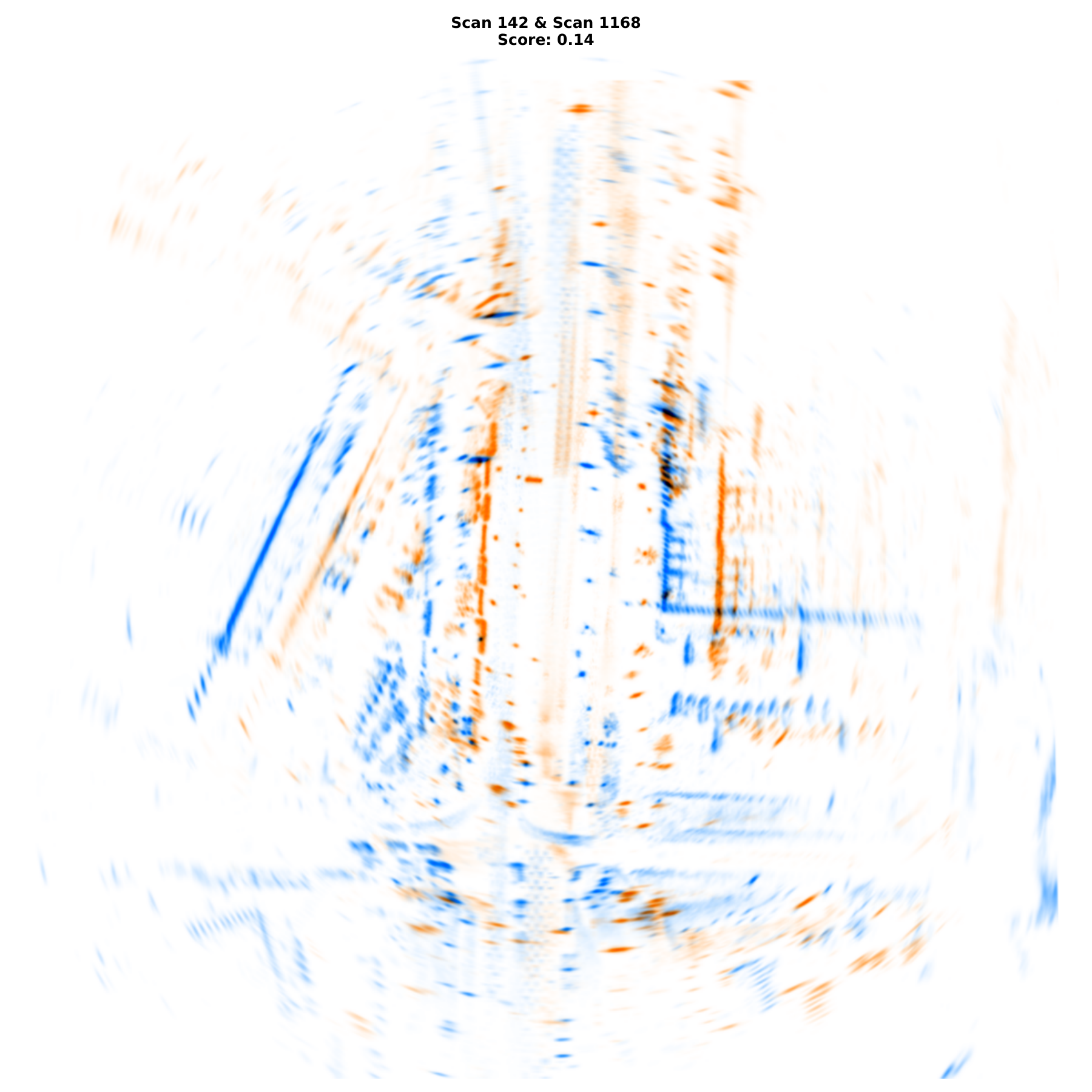}
    \end{minipage}%
    \begin{minipage}[t]{0.48\linewidth}
        \centering
        \includegraphics[clip, trim=\trimsizeB cm \trimsizeB cm \trimsizeB cm \trimsizeB cm,width=\imgwidth\linewidth]{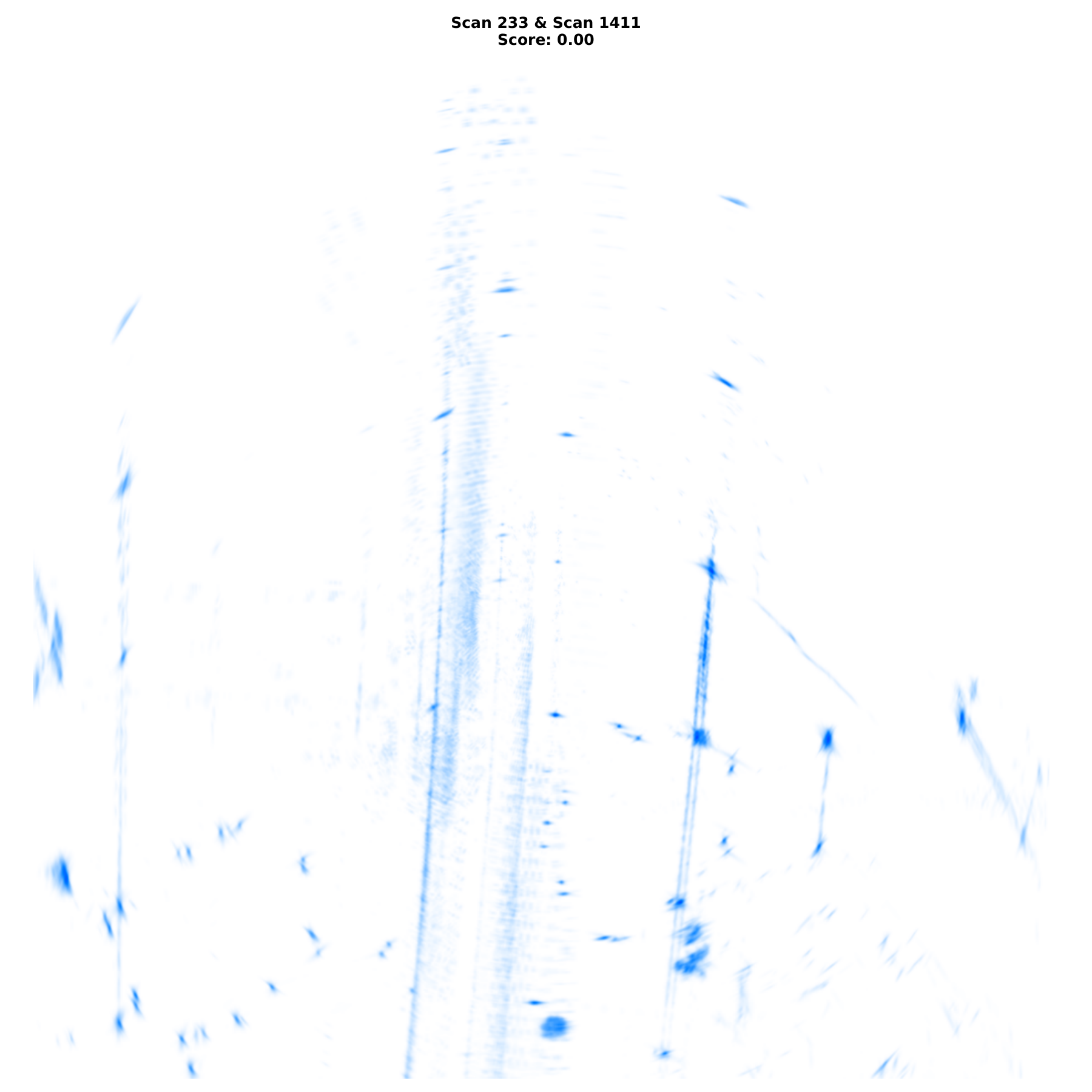}
    \end{minipage}%
    \\
    [\rowspace]
    
    \begin{minipage}[t]{0.48\linewidth}
        \centering
        \includegraphics[clip, trim=\trimsize cm \trimsize cm \trimsize cm \trimsize cm,width=\imgwidth\linewidth]{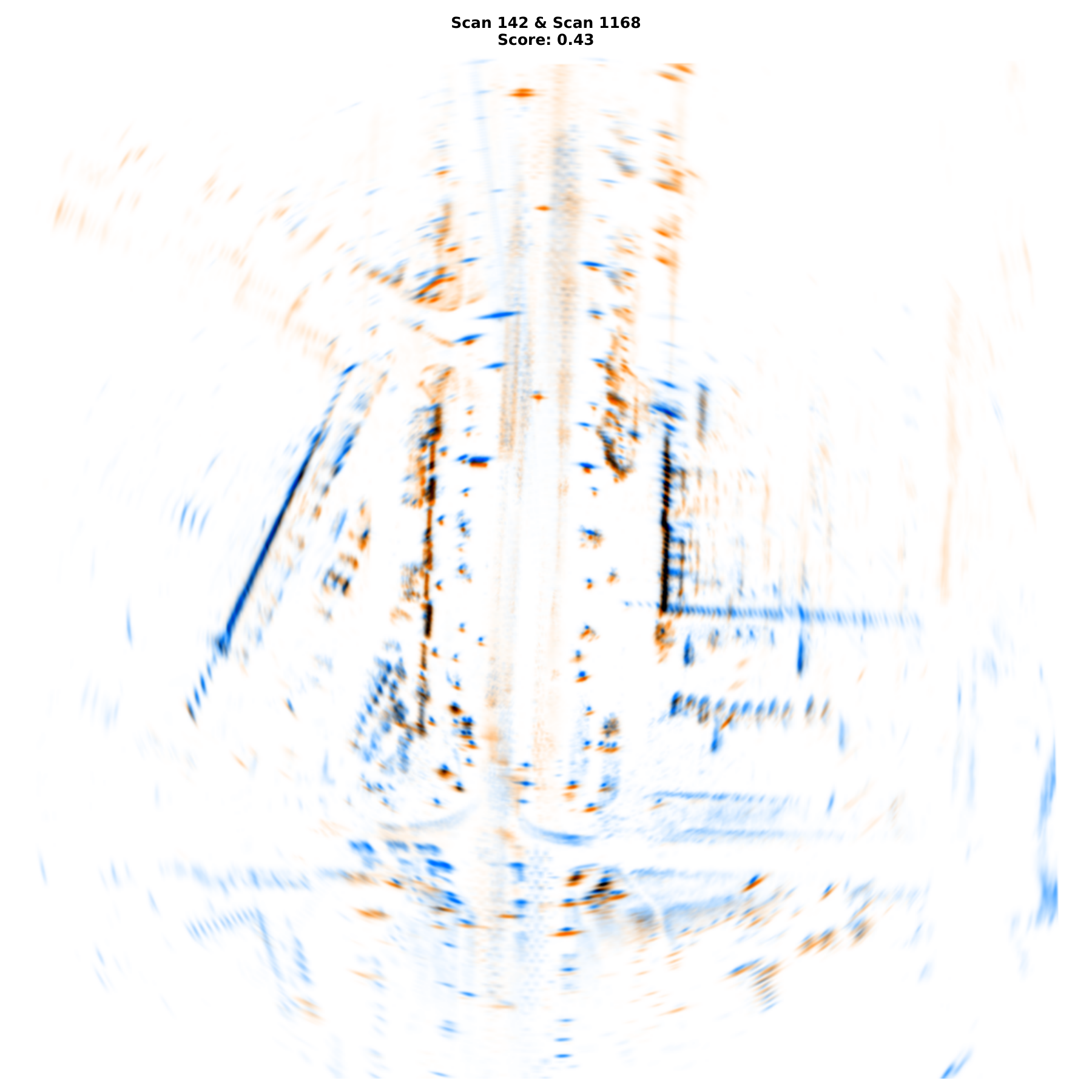}
    \end{minipage}%
    \begin{minipage}[t]{0.48\linewidth}
        \centering
        \includegraphics[clip, trim=\trimsizeB cm \trimsizeB cm \trimsizeB cm \trimsizeB cm,width=\imgwidth\linewidth]{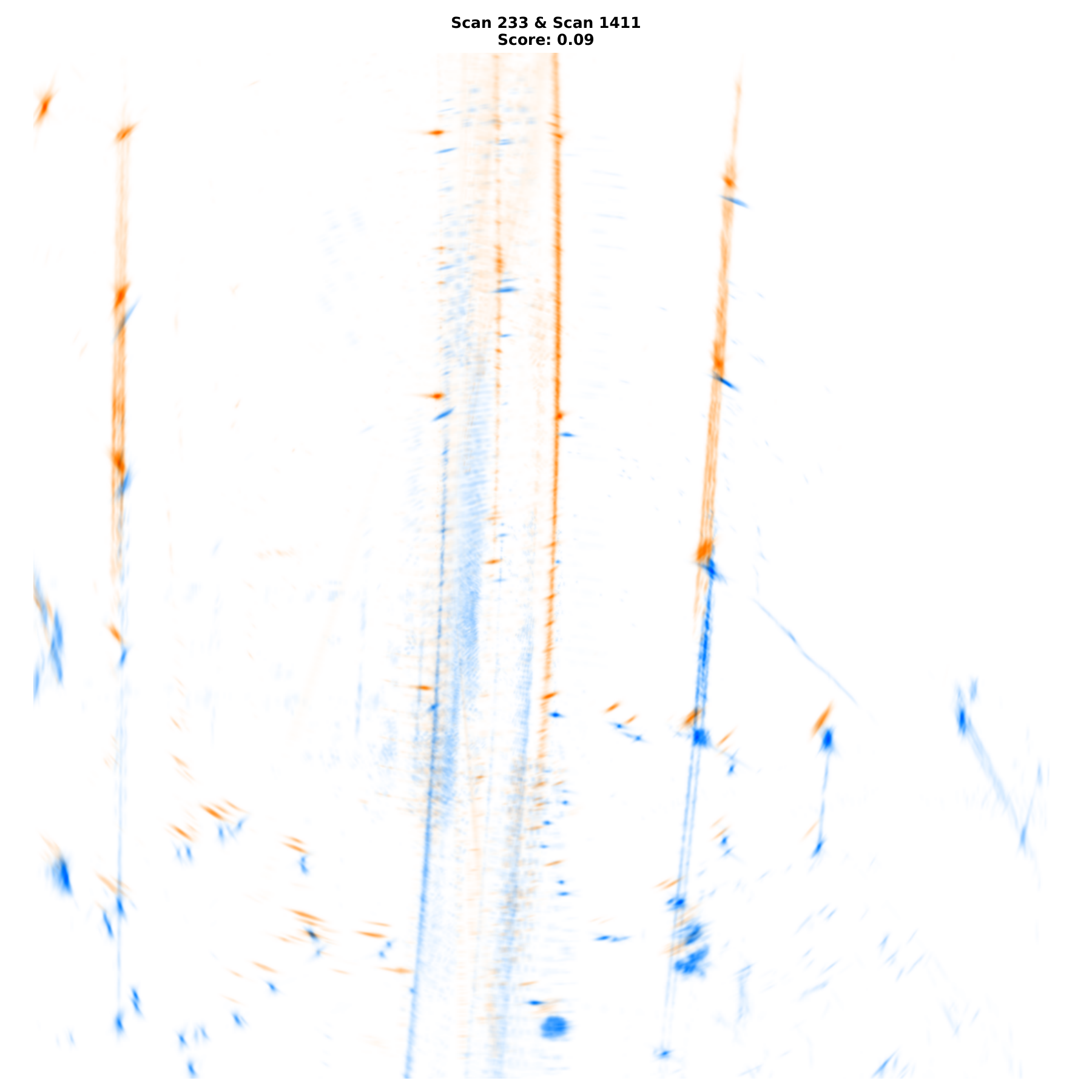}
    \end{minipage}%
    \\
    [\rowspace]
    \begin{minipage}[t]{0.48\linewidth}
        \centering
        \includegraphics[clip, trim=\trimsize cm \trimsize cm \trimsize cm \trimsize cm, width=\imgwidth\linewidth]{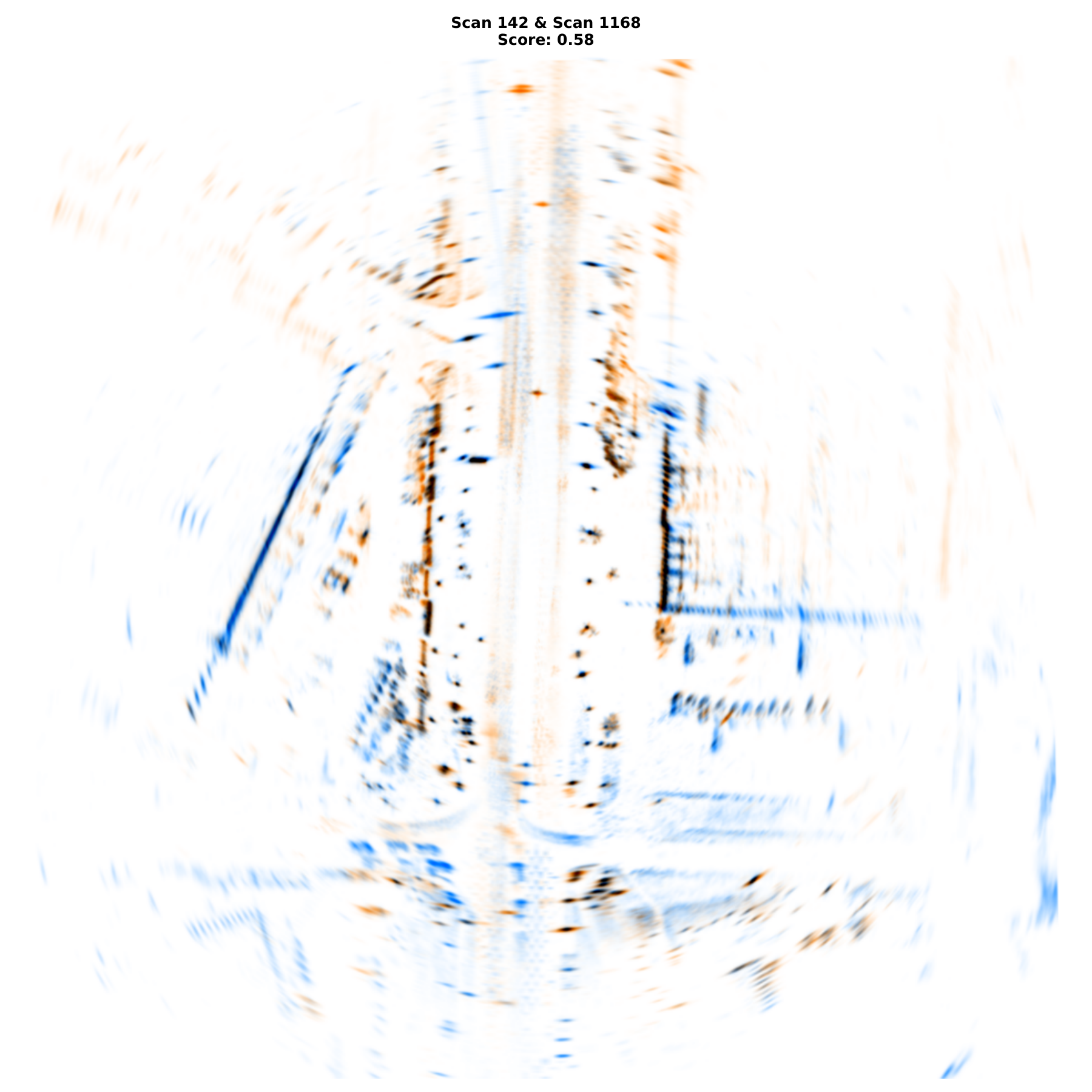}
    \end{minipage}%
    \begin{minipage}[t]{0.48\linewidth}
        \centering
        \includegraphics[clip, trim=\trimsizeB cm \trimsizeB cm \trimsizeB cm \trimsizeB cm, width=\imgwidth\linewidth]{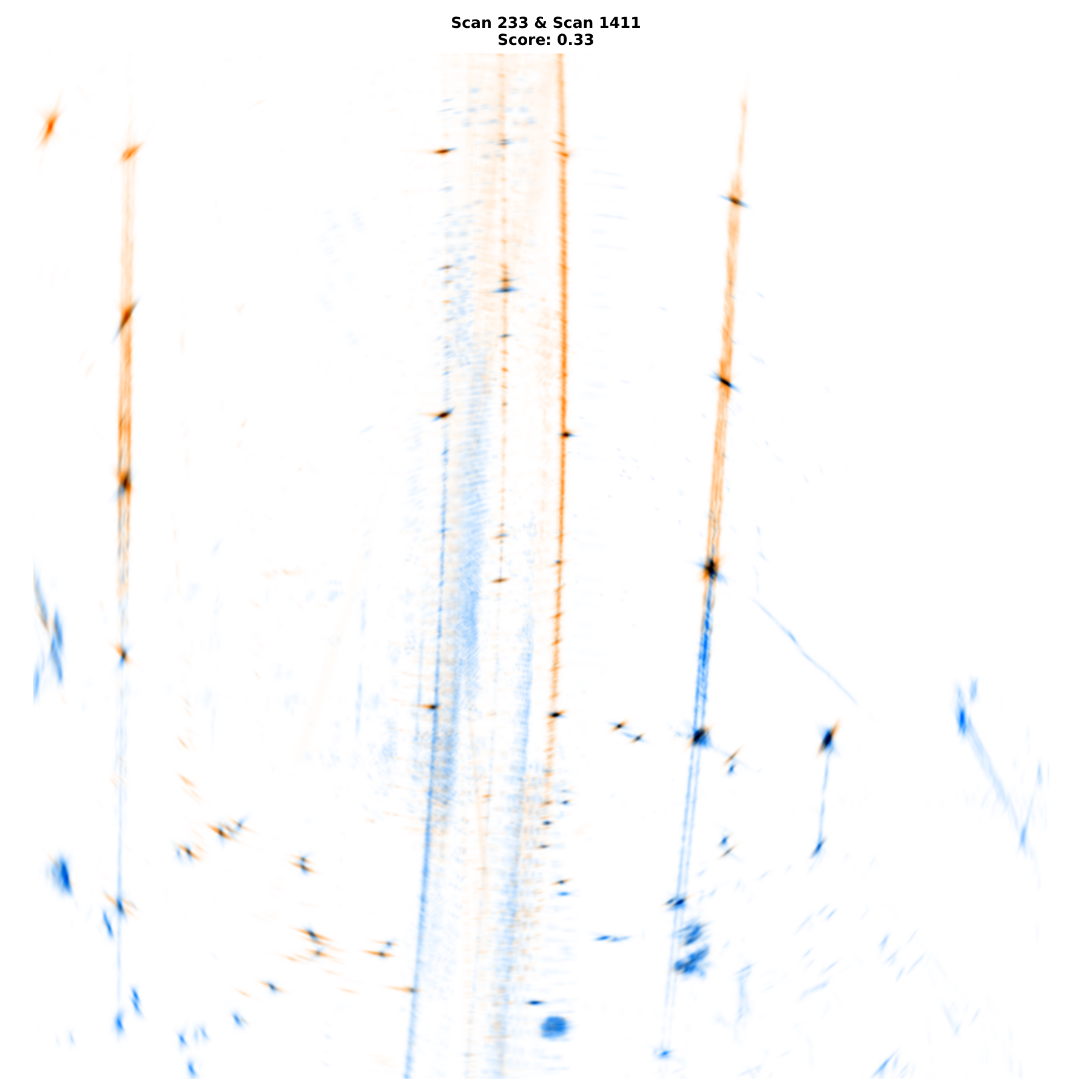}
    \end{minipage}%
    \end{minipage}
    }
    };

    \node[anchor=center] at ($(fig.north west)+(2.3cm,-11.1cm)$) {\texttt{Suburbs}};
    \node[anchor=center] at ($(fig.north west)+(6.3cm,-11.1cm)$) {\texttt{Skyway}};

    \node[anchor=center] at ($(fig.north west)+(0.4cm,-1.9cm)$) {\rotatebox{90}{TBV-SLAM \cite{adolfsson2023tbv}}};
    
    \node[anchor=center] at ($(fig.north west)+(0.4cm,-5.4cm)$) {\rotatebox{90}{Dr-PoGO init. \cite{legentil2026drpogo}}};
    
    \node[rotate=90, anchor=center] 
      at ($(fig.north west)+( 0.4cm,-8.9cm)$) {Dr-BA (ours)};

    \node[
        anchor=north,
        text width=4.2cm,
        align=center
    ] at ($(fig.north west)+(6.3cm,-0.1cm)$)
    {%
    \textcolor{red}{trajectory divergence}\\
    {\scriptsize corresponding scan estimated \SI{1214}{\meter} away}%
    };
    
    \end{tikzpicture}
    
    \caption{Qualitative comparison of trajectory estimates from TBV-SLAM, our pose-graph initialization (Dr-PoGO), and the proposed bundle adjustment method Dr-BA. Each image overlays two scans from the same location collected in opposite driving directions (blue/orange). While Dr-PoGO achieves low ATE, it does not guarantee local consistency (see longitudinal misalignment). Dr-BA uniquely enforces local consistency, producing well-overlapping scans critical for downstream tasks such as localization.}
    \label{fig:map_comparison}
\end{figure}

\subsection{Metrics}
The \ac{ate} evaluates the overall accuracy of Dr-BA and the baseline methods for trajectory estimation.
After aligning the estimated and ground-truth trajectories via SVD-based $SE(2)$ registration, the \ac{ate} is computed as $\sqrt{\frac{1}{N}\sum_{n=1}^{N}\Vert\mbfhat{r}_n - \mbf{r}_n^{\gt}\Vert^2}$, where $\mbfhat{r}_n$ is the translation component of the pose estimate $\mbfhat{T}_n$ and $\mbf{r}_n^{\gt}$ is the ground-truth translation.
Since all sequences finish close to the starting location, we use the \ac{epe} to assess the effectiveness of loop-closure constraints in each method.
The \ac{epe} is defined as the norm of the translation component of $(\mbf{T}_N^{\gt})^{-1}\mbf{T}_1^{\gt}(\mbfhat{T}_n)^{-1}\mbfhat{T}_N$.
We also introduce a `self-consistency' metric to evaluate the quality of the produced maps.
For each mapping pose, we identify the nearest pose that is at least $300\,\si{\meter}$ away by distance travelled, but within $25\,\si{\meter}$ Euclidean distance, then compute the relative error between the associated poses in the estimated trajectory and the ground truth. This metric can be interpreted as a localization error within a trajectory when revisiting previously explored areas. The more accurately these poses align, the more locally consistent the resulting map is. 
To assess the accuracy of the proposed localization pipeline, which further validates the mapping accuracy, we analyze the relative pose error between the trajectory used for mapping and the localization run following the Boreas dataset localization benchmark~\cite{burnett2023boreas}.

\begin{table}[t]
    \centering
    \caption{Average translation ([\si{\m}]) \& rotation ([\si{\degree}]) self-consistency RMSE per sequence type for Dr-PoGO and Dr-BA.}
    \setlength{\tabcolsep}{2pt}
    \small
    \begin{tabularx}{\linewidth}{lYY}
        \toprule
        \textbf{Sequence} & \textbf{Dr-PoGO} \cite{legentil2026drpogo} & \textbf{Dr-BA (ours)}
        \\
        \midrule
        \texttt{Suburbs}    & 0.34 / 0.11 & \textbf{0.14} / \textbf{0.09} \\
        \texttt{Industrial} & 0.36 / 0.18 & \textbf{0.10} / \textbf{0.10} \\
        \texttt{Skyway}     & 3.49 / \textbf{0.10} & \textbf{1.29} / 0.24 \\
        \texttt{Forest}     & 0.57 / \textbf{0.12} & \textbf{0.52} / \textbf{0.12} \\
        \texttt{Farm}       & 0.28 / \textbf{0.08} & \textbf{0.16} / 0.11 \\
        \bottomrule
        \multicolumn{3}{l}{\scriptsize XX / YY, with XX the translation and YY the rotation RMSE.}
    \end{tabularx}
    \label{tab:self_consistency_eval}
\end{table}

\subsection{Direct radar bundle adjustment}
Dr-BA is compared to three baselines.
First, we compare to a \ac{sota} odometry method, DRO \cite{legentil2025dro}, with gyroscope measurements and explicit Doppler compensation, with no loop closures.
This approach, unsurprisingly, performs the worst.
Second, we compare to an off-the-shelf \ac{sota} radar \ac{slam} method TBV-SLAM \cite{adolfsson2023tbv}, which employs a robust loop closure candidate verification approach.
Finally, we evaluate performance using Dr-PoGO \cite{legentil2026drpogo}.
Table~\ref{tab:traj_eval} shows the average \ac{ate} and \ac{epe} metrics for all considered methods.
Figure~\ref{fig:teaser} shows a full Dr-BA-created map of an \texttt{Industrial} sequence.
Table~\ref{tab:self_consistency_eval} shows the translation and rotation self-consistency RMSE from Dr-PoGO and Dr-BA for all sequences in all routes. We omit DRO and TBV-SLAM results for brevity and because they were significantly worse ($>10\,\si{\meter}$).

We note that Dr-PoGO already achieves notably better performance than TBV-SLAM, likely due to the high-quality odometry provided by DRO.
Dr-PoGO improves \ac{ate} performance by three to ten times on the structured \texttt{Suburbs} and \texttt{Industrial} sequences, and prevents failure on the challenging sequences.
In environments with extensive trajectory overlap (\texttt{Suburbs}, \texttt{Skyway}) or well-defined geometric features (\texttt{Industrial}), Dr-BA further refines the Dr-PoGO trajectory, achieving the best average \ac{ate} and \ac{epe} results.
In the unstructured environments with very little trajectory overlap (\texttt{Forest}, \texttt{Farm}), Dr-BA performs on par with Dr-PoGO.
However, the most significant improvements can be found in the local consistency of the produced maps.
Figure~\ref{fig:map_comparison} highlights regions from a \texttt{Suburbs} and \texttt{Skyway} sequence where Dr-PoGO yields a good trajectory, but lacks local map consistency.
In contrast, Dr-BA generates a locally consistent map.
The self-consistency metric further backs up this finding.
Dr-BA significantly improves translational self-consistency on all routes with notable pose overlap (meaning all except \texttt{Forest}). Rotational self-consistency stays roughly similar, except for the \texttt{Skyway}, where the adaptive blurring results in a loss of rotational self-consistency in exchange for a significant boost in translational performance.

In practice, we find that $r_v = \SI{1.0}{\m}$ is more than sufficient for high quality \ac{ba}.
This choice also helps maintain a multi-threaded runtime of half an hour to a few hours per map on a standard laptop, depending on the trajectory length and parameters, with maps being represented by up to four million points.
Further runtime improvements, such as GPU-acceleration, are left for future work.

\begin{table}[t]
    \centering
    \caption{RMSE localization errors for RTR and DRL. DRL is run on maps created using Dr-BA, ground truth (GT), and Dr-PoGO poses. \texttt{Forest} results are greyed out as they are at the limit of groundtruth noise.}
    \setlength{\tabcolsep}{2pt}
    \begin{tabularx}{\linewidth}{lYYY}
        \toprule
        \textbf{Seq. type \& Method} & \textbf{Long.} \scriptsize[\si{\m}] & \textbf{Lat.} \scriptsize[\si{\m}] & \textbf{Yaw} \scriptsize[$^\circ$] 
        \\
        \midrule
        \texttt{Suburbs} &
        \\\quad RTR \scriptsize\cite{are_we_ready_for} & 0.092 & 0.052 & 0.107
        \\\quad DRL-Dr-BA \scriptsize(ours) & 0.076 & \textbf{0.049} & \textbf{0.061}
        \\\quad DRL-GT & \textbf{0.071} & 0.061 & 0.074
        \\\quad DRL-Dr-PoGO & 0.146 & 0.105 & 0.078
        \\
        \midrule
        \texttt{Industrial} &
        \\\quad RTR \scriptsize\cite{are_we_ready_for} & 0.080 & \textbf{0.054} & 0.116
        \\\quad DRL-Dr-BA \scriptsize(ours) & 0.081 & 0.062 & \textbf{0.055}
        \\\quad DRL-GT & \textbf{0.076} & 0.082 & 0.073
        \\\quad DRL-Dr-PoGO & 0.109 & 0.076 & 0.085
        \\
        \midrule
        \texttt{Skyway} &
        \\\quad RTR \scriptsize\cite{are_we_ready_for} & - & - & -$^3$
        \\\quad DRL-Dr-BA \scriptsize(ours) & 0.301 & 0.200 & 0.145 
        \\\quad DRL-GT & \textbf{0.220} & \textbf{0.126} & \textbf{0.090}
        \\\quad DRL-Dr-PoGO & 1.517 & 0.419 & 0.157
        \\
        \midrule
        \texttt{Forest} &
        \\\quad RTR \scriptsize\cite{are_we_ready_for} & - & - & -$^3$
        \\\quad DRL-Dr-BA \scriptsize(ours) & \textcolor{gray}{0.429} & \textcolor{gray}{0.371} & \textcolor{gray}{0.122}
        \\\quad DRL-GT & \textcolor{gray}{0.415} & \textcolor{gray}{0.369} & \textcolor{gray}{0.138}
        \\\quad DRL-Dr-PoGO & \textcolor{gray}{0.425} & \textcolor{gray}{0.371} & \textcolor{gray}{0.141}
        \\
        \midrule
        \texttt{Farm} &
        \\\quad RTR \scriptsize\cite{are_we_ready_for} & - & - & -$^3$
        \\\quad DRL-Dr-BA \scriptsize(ours) & 0.404 & 0.595 & 0.285
        \\\quad DRL-GT & \textbf{0.155} & \textbf{0.159} & \textbf{0.107}$^1$
        \\\quad DRL-Dr-PoGO & 0.191 & 0.251 & 0.148$^2$
        \\
        \bottomrule
    \\[-2.0ex]
    \multicolumn{4}{l}{\scriptsize $^1$ The superscript indicates the number of failed sequences.}
    \end{tabularx}
    \label{tab:localization}
\end{table}

\subsection{Direct radar localization}
While trajectory and visual evaluations are useful for comparing \ac{ba} algorithms, their ultimate goal is to produce high-quality maps for downstream tasks.
To assess the practical impact of Dr-BA, we evaluate its maps within our novel direct radar localization framework, DRL.
We note that these maps are also compatible with cross-correlation-based localization or point-cloud methods by representing non-zero map regions as points.

For a full comparison, we construct maps using \eqref{eq:i_as_func_pose} with poses from Dr-BA (DRL-Dr-BA), ground truth (DRL-GT), and Dr-PoGO (DRL-Dr-PoGO).
DRL localizes a single scan to the map, using DRO to propagate the estimate to the next frame, without explicit priors or costs beyond the direct intensity objective in \eqref{eq:loc_objective}.
We again find that $r_v = \SI{1.0}{\m}$ is sufficient for high-quality localization.

We compare DRL with an open-source point cloud-based radar localization framework, radar teach \& repeat (RTR) \cite{are_we_ready_for}.
RTR creates locally consistent, but globally inconsistent, point maps of the environment, and then does topometric localization relative to them on repeated traversals of the same environment.

Table~\ref{tab:localization} presents the longitudinal, lateral, and yaw localization error RMSE for all sequence types and methods.
On the structured \texttt{Suburbs} and \texttt{Industrial} routes, DRL-Dr-BA and DRL-GT achieve \ac{sota} performance, keeping localization errors within $10\,\si{\cm}$ and $0.1\si{\degree}$.
While translational performance is comparable to RTR, DRL-Dr-BA substantially outperforms RTR, and slightly outperforms DRL-GT, in rotation.
This is likely due to higher local map quality, as even small rotational inconsistencies at the radar frame cause large intensity offsets at long ranges.
The value of Dr-BA is further shown by the fact that DRL-Dr-BA outperforms DRL-Dr-PoGO on all metrics on the \texttt{Suburbs}, \texttt{Industrial}, and \texttt{Skyway} sequences.

Performance for all methods drops on the more challenging sequences, with RTR failing to localize until the end of all \texttt{Skyway}, \texttt{Forest}, and \texttt{Farm} sequences.
In contrast, DRL-based methods only experience failures on \texttt{Farm} sequences, with DRL-Dr-BA not having any unrecoverable failures across the entire dataset, something not achieved even by DRL-GT.
Despite preventing failures, the Dr-BA map yields worse average localization performance in the successful \texttt{Farm} sequences than DRL-GT or DRL-Dr-PoGO.
Possible reasons for this are outlined in Section~\ref{sec:limitations}.
All DRL methods produce similar results on the \texttt{Forest} route, as performance is limited by groundtruth noise due to the large distance to the nearest RTK base station and tree canopy \cite{lisus_brrt26}.
Nearly identical errors are reported by baselines in \cite{lisus_brrt26} and in the \ac{sota} lidar method in \cite{legentil20242fast2lamaa}.
These results are included for completeness but not compared between implementations.

DRL takes on average $200\,\si{\ms}$ per frame running on a single thread of a Lenovo P1 Laptop with Intel(R) Core(TM) i7-12800H CPU @ $4.8 \, \si{\giga\hertz}$ and $32 \, \si{\giga\byte}$ of RAM, making it `real-time' for the $4\,\si{\hertz}$ radar used in these tests.

\begin{table}[]
    \centering
    \caption{Ablation study on the impact of different parameters on \ac{ba} and consequently localization performance. Average ATE/EPE/localization translation RMSE per sequence type are shown (all [\si{\m}]). The `Dr-BA' method uses cumulative images (cumul. img.), adaptive blurring (AdBlur), and local maps (LocMaps.)} 
    \setlength{\tabcolsep}{2pt}
    \small
    \begin{tabularx}{\linewidth}{lYYY}
        \toprule
        \textbf{Method} & \texttt{Suburbs} & \texttt{Skyway}
        \\
        \midrule
         Dr-BA & \textbf{0.51}/0.15/\textbf{0.09} &  \textbf{1.33}/0.10/\textbf{0.36}
         \\
         AdBlur: off & 0.53/0.15/\textbf{0.09} &  1.60/0.10/1.19
         \\
         No cumul. img. &  0.54/0.20/0.10  &  \textbf{1.33}/0.09/0.39
        \\
        No LocMaps & 0.54/\textbf{0.13}/0.10 & 1.73/\textbf{0.08}/1.37$^2$
        \\
        \bottomrule
        \multicolumn{3}{l}{\scriptsize $^1$ The superscript indicates the number of failed sequences.}
    \end{tabularx}
    \label{tab:ba_ablation}
\end{table}

\begin{figure}[t]
    \centering
    \includegraphics[width=\linewidth]{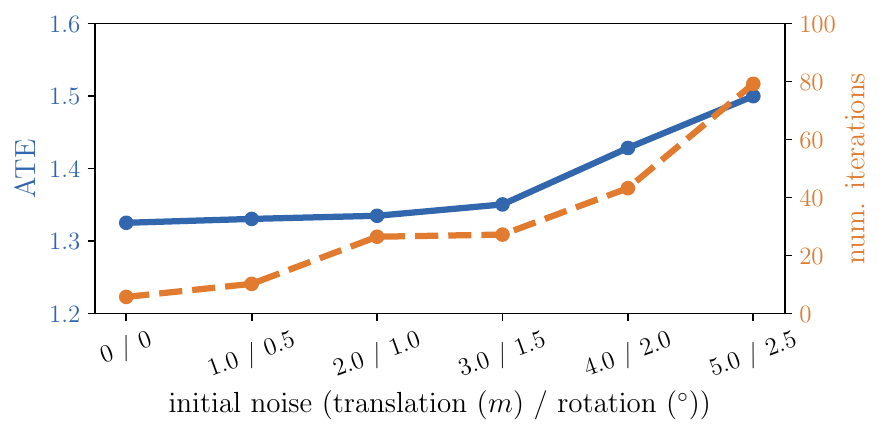}
    \caption{Average ATE and convergence for the \texttt{Industrial} route with increasingly degraded Dr-PoGO initialization.}
    \label{fig:init_sensitivity}
\end{figure}

\subsection{Ablation studies}
\label{sec:ablation}
\subsubsection{Processing impact}
Table~\ref{tab:ba_ablation} presents an ablation study on the impact of different processing parameters on \ac{ba} and ultimately localization performance.
Ablations are done on only a \texttt{Suburbs} and \texttt{Skyway} sequence as they contain a large number of overlapping regions in their traversals.
DRL-Dr-BA localization results on the remaining three sequences are also included, using each map from the ablation study, for a complete evaluation.
All options generally perform well on the \texttt{Suburbs} route, though combining all options yields a slight improvement in \ac{ate}.
This is due to the \texttt{Suburbs} having consistently well-defined geometric structure in all scans.
However, the use of adaptive blurring and local maps is required to get good performance on the \texttt{Skyway} route, where the radar has very little to observe, as shown in Figure~\ref{fig:challenging_segments} (a).

\subsubsection{Initialization sensitivity}
We add increasingly aggressive random (uniform) initialization error to Dr-BA's PGO initialization and plot ATE and iterations to convergence in Figure~\ref{fig:init_sensitivity}.
Results show average performance from all \texttt{Industrial} sequences (chosen for smaller size and faster testing).
Perturbations up to $3.0\,\si{\meter}$ and $1.50\si{\degree}$ are recovered without significant ATE degradation, although convergence naturally takes longer.
This demonstrates that while general global consistency is required to initialize Dr-BA, the local consistency can be quite poor.

\subsubsection{VarPro impact}
\begin{figure}[t]
    \centering
    \includegraphics[width=\linewidth]{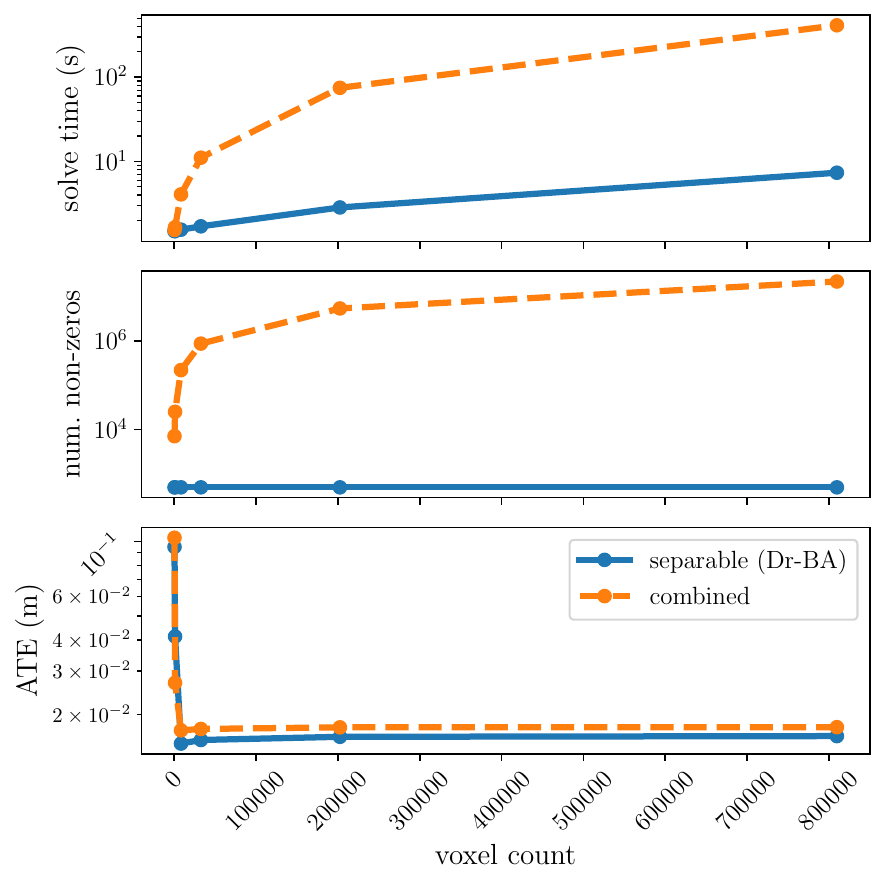}
    \caption{Total solve time, Hessian memory use, and ATE comparison between separable and combined Dr-BA.}
    \label{fig:timing_tests}
\end{figure}

To demonstrate the benefits of \ac{varpro} over the naive solution, we construct a test problem with 11 poses (first fixed) at a stationary location and progressively increase voxel resolution to scale the problem size.
We evaluate \ac{ate}, solve time, and Hessian non-zeros (as a memory proxy).
Figure~\ref{fig:timing_tests} compares our separable Dr-BA implementation against a combined implementation that jointly solves \eqref{eq:full_ba_objective} for poses and voxel intensities.
Both yield near-identical ATE, but the combined formulation requires substantially more compute and memory, becoming prohibitively slow for full sequences.

\subsection{Limitations}
\label{sec:limitations}
\begin{figure}
    \centering
    \begin{subfigure}[t]{\linewidth}
        \centering
        \includegraphics[width=\linewidth, clip, trim=0cm 14cm 0cm 14cm]{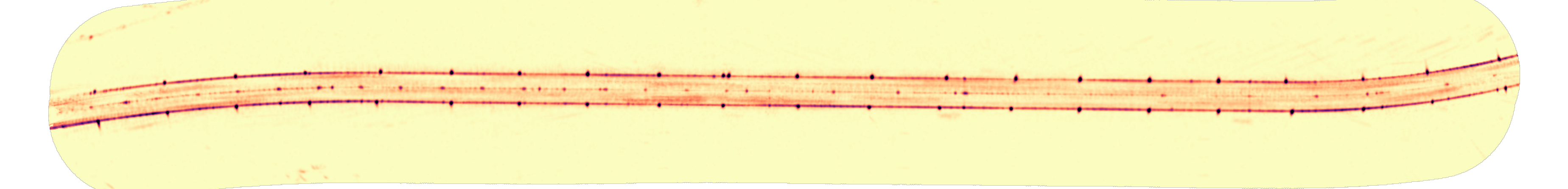}
        \caption{\texttt{Skyway} segment}
    \end{subfigure}
    \begin{subfigure}[t]{\linewidth}
        \centering
        \includegraphics[width=\linewidth, clip, trim=0cm 9cm 0cm 14cm]{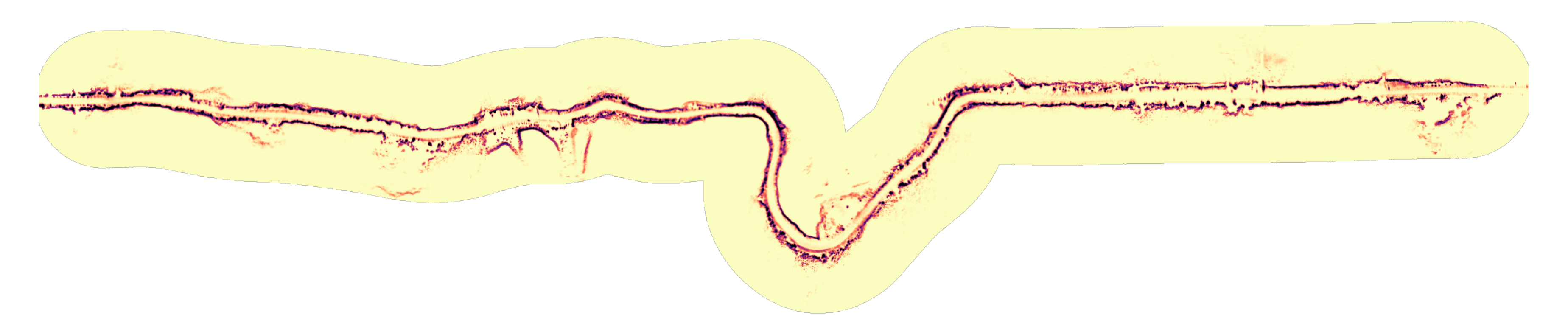}
        \caption{\texttt{Forest} segment}
    \end{subfigure}
    \begin{subfigure}[t]{\linewidth}
        \centering
        \includegraphics[width=\linewidth, clip, trim=0cm 12cm 0cm 14cm]{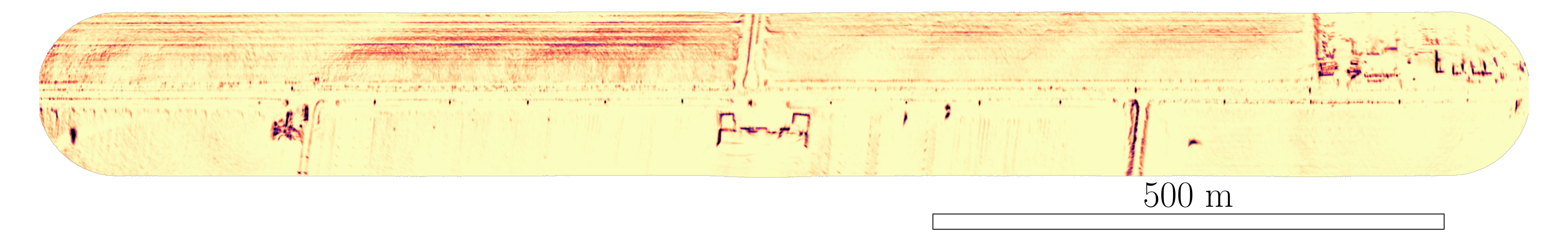}
        \caption{\texttt{Farm} segment}
    \end{subfigure}

    \caption{Challenging segments of the \texttt{Skyway}, \texttt{Farm}, and \texttt{Forest} routes, characterized by sparse features, diffuse natural structure, or poor scan co-visibility. The lengthscale is shared by all plots.}
    \label{fig:challenging_segments}
\end{figure}

Despite \ac{sota} performance on structured sequences and a decreased failure rate in challenging unstructured environments, there is a notable decrease in performance of Dr-BA and DRL on the challenging routes.
Particularly challenging segments of these routes are shown in Figure~\ref{fig:challenging_segments}, where all three difficult sequences contain extensive regions with very few features visible from a single radar scan.
The \texttt{Forest} sequences also contain tight vegetation corridors with poor co-visibility from scans collected short distances apart.
These are fundamental sensor limitations in a given environment, and additional interoceptive data must be included in the estimation to mitigate them.
This is likely why DRL-Dr-PoGO performs better in successful localization runs than DRL on the \texttt{Farm} sequences, as Dr-PoGO poses, and consequently the map produced using them, are constrained by relative motion information provided by Doppler velocity measurements and a gyroscope.
Additionally, this work ignores the dependency of radar intensity returns on the viewpoint from which the sensor is observing a feature.
Explicitly incorporating the viewpoint significantly increases complexity, but would likely yield a more accurate \ac{ba} result.

\section{Conclusion}
This paper introduces Dr-BA, a first-of-its-kind direct radar bundle adjustment framework.
Dr-BA is based on a general separable formulation that enables independent optimization of the trajectory and the map, resulting in an efficient algorithm with linear scaling in the number of estimated map states.
Using 2D spinning radar data, we demonstrate that Dr-BA can tractably construct dense maps spanning tens of kilometres while maintaining both local and global consistency.
Building on this formulation, we also propose a novel direct radar localization (DRL) method.
Experimental results on over $200\,\si{\km}$ of real-world data show \ac{sota} performance in both \ac{ba} and localization using Dr-BA-generated maps.
Tests are run on five diverse trajectories, ranging from structured suburban environments to highly unstructured farm roads.


\bibliographystyle{plainnat}
\bibliography{references}

\newpage
\clearpage
\onecolumn

\section*{Supplementary Material}

\setcounter{section}{0}
\renewcommand{\thesection}{\Alph{section}}
\setcounter{figure}{0}
\renewcommand{\thefigure}{\thesection.\arabic{figure}}

\section{Detailed Sequence Visuals}
\subsection{Suburbs}

\begin{figure}[htbp]
    \centering
    \begin{tikzpicture}[spy using outlines={magnification=3, size=3cm, connect spies, dashed, ultra thick}]
        \node[inner sep=0, anchor=north] (map) at (0.0,0.0)
            {\includegraphics[width=\linewidth,
            clip, trim=158mm 0mm 158mm 00mm]{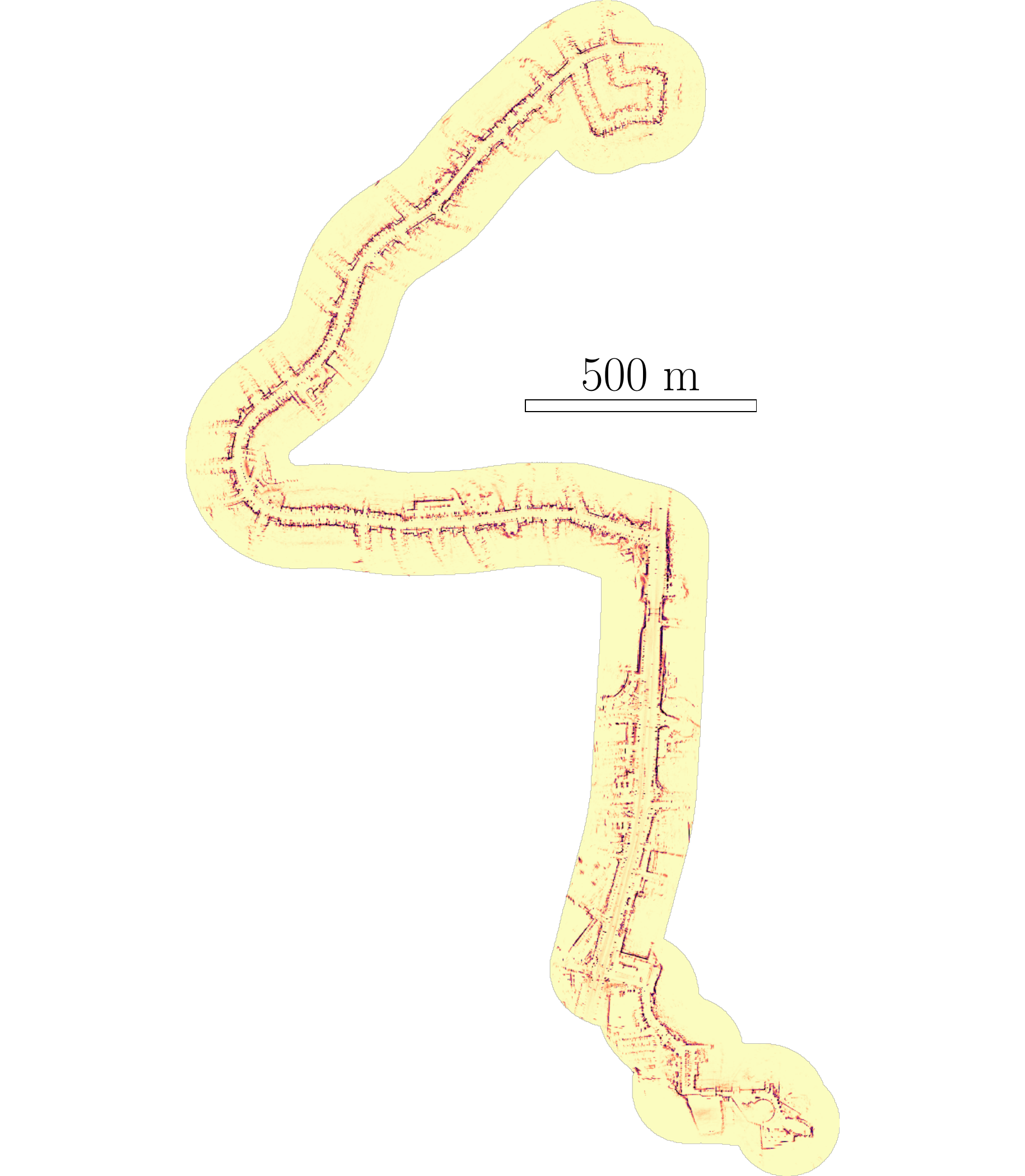}};

        \node at (-4.5,-16) {
          \begin{minipage}{0.4\linewidth}
            \centering
            \includegraphics[width=\linewidth]{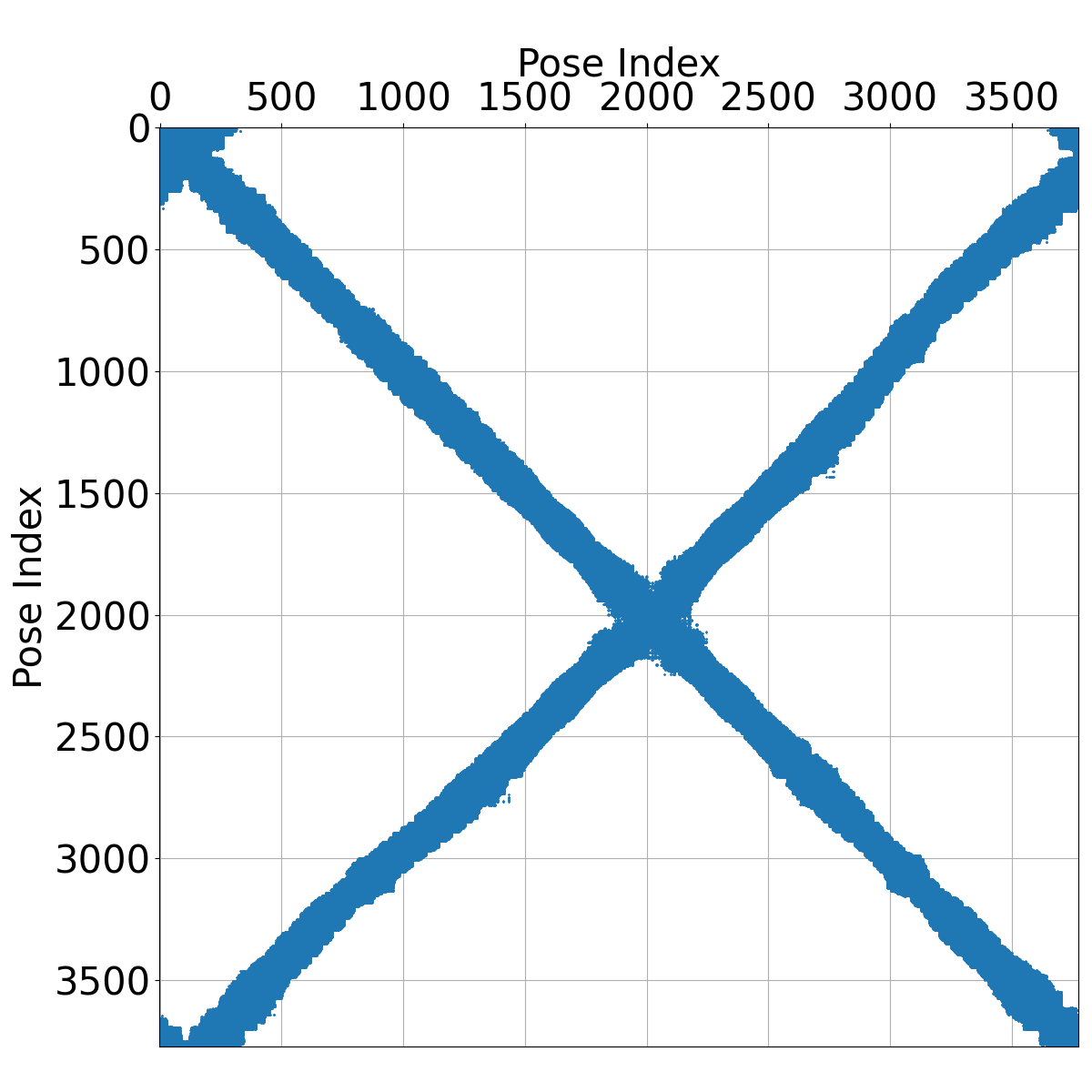}
            \captionof{figure}{Sparsity pattern of $\mbf{H}$ (from \eqref{eq:linear_problem}) for a \texttt{Industrial} sequence.}
          \end{minipage}
        };
        
    \end{tikzpicture}
\end{figure}

\clearpage
\subsection{Industrial}

\begin{figure}[htbp]
    \centering
    \begin{tikzpicture}
        \node[inner sep=0, anchor=north] (map) at (0.0,0.0)
            {\includegraphics[width=\linewidth,
            clip, trim=100mm 0mm 100mm 00mm]{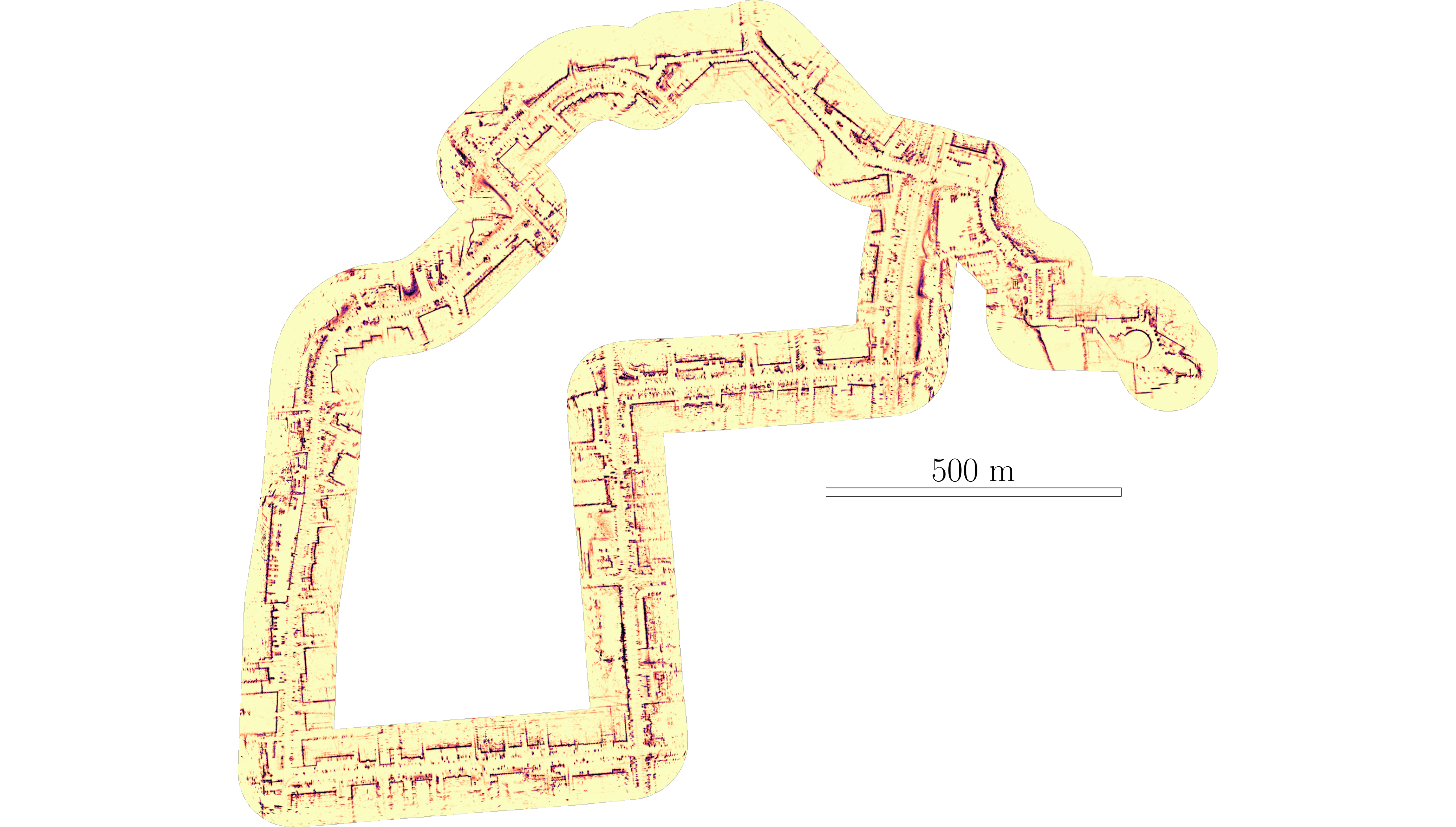}};

        \node at (4.0,-16) {
          \begin{minipage}{0.4\linewidth}
            \centering
            \includegraphics[width=\linewidth]{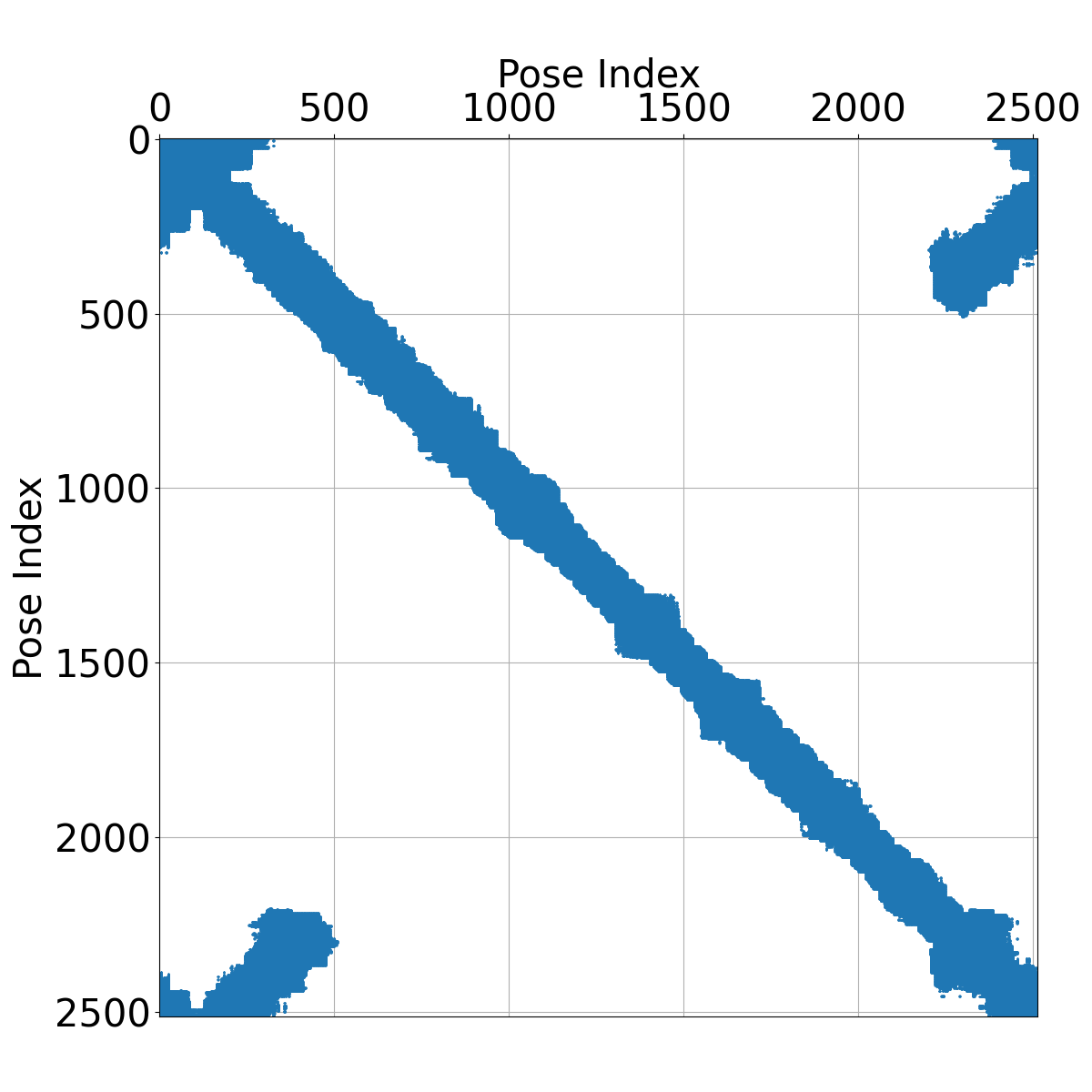}
            \captionof{figure}{Sparsity pattern of $\mbf{H}$ (from \eqref{eq:linear_problem}) for a \texttt{Industrial} sequence.}
          \end{minipage}
        };
        
    \end{tikzpicture}
\end{figure}

\clearpage
\subsection{Skyway}

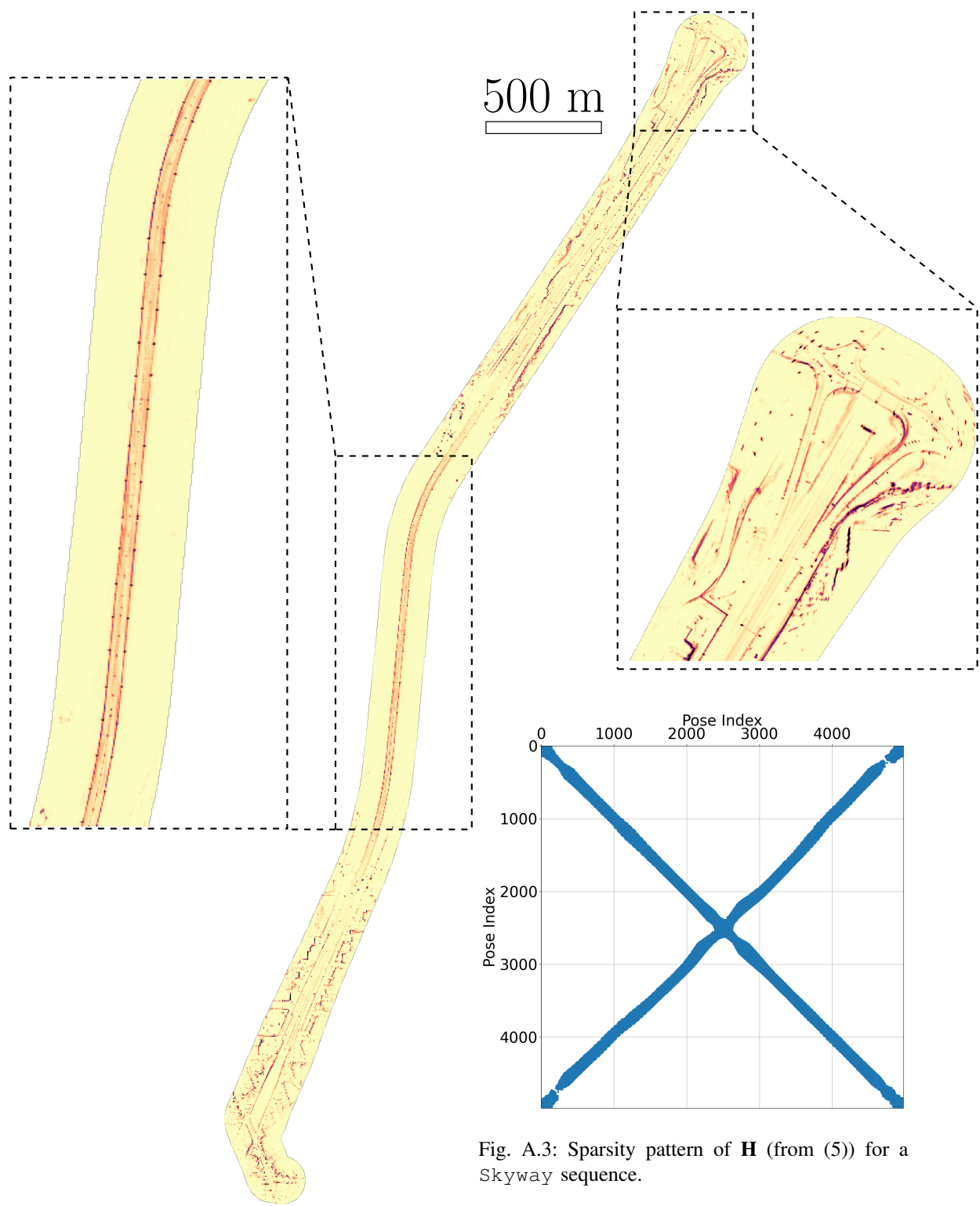
\begin{figure}[htbp]
    \centering
    
    \begin{tikzpicture}
    
        \node[inner sep=0, anchor=north] (map) at (0.0,0.0)
            {\includegraphics[width=\linewidth,
            clip, trim=155mm 0mm 155mm 00mm]{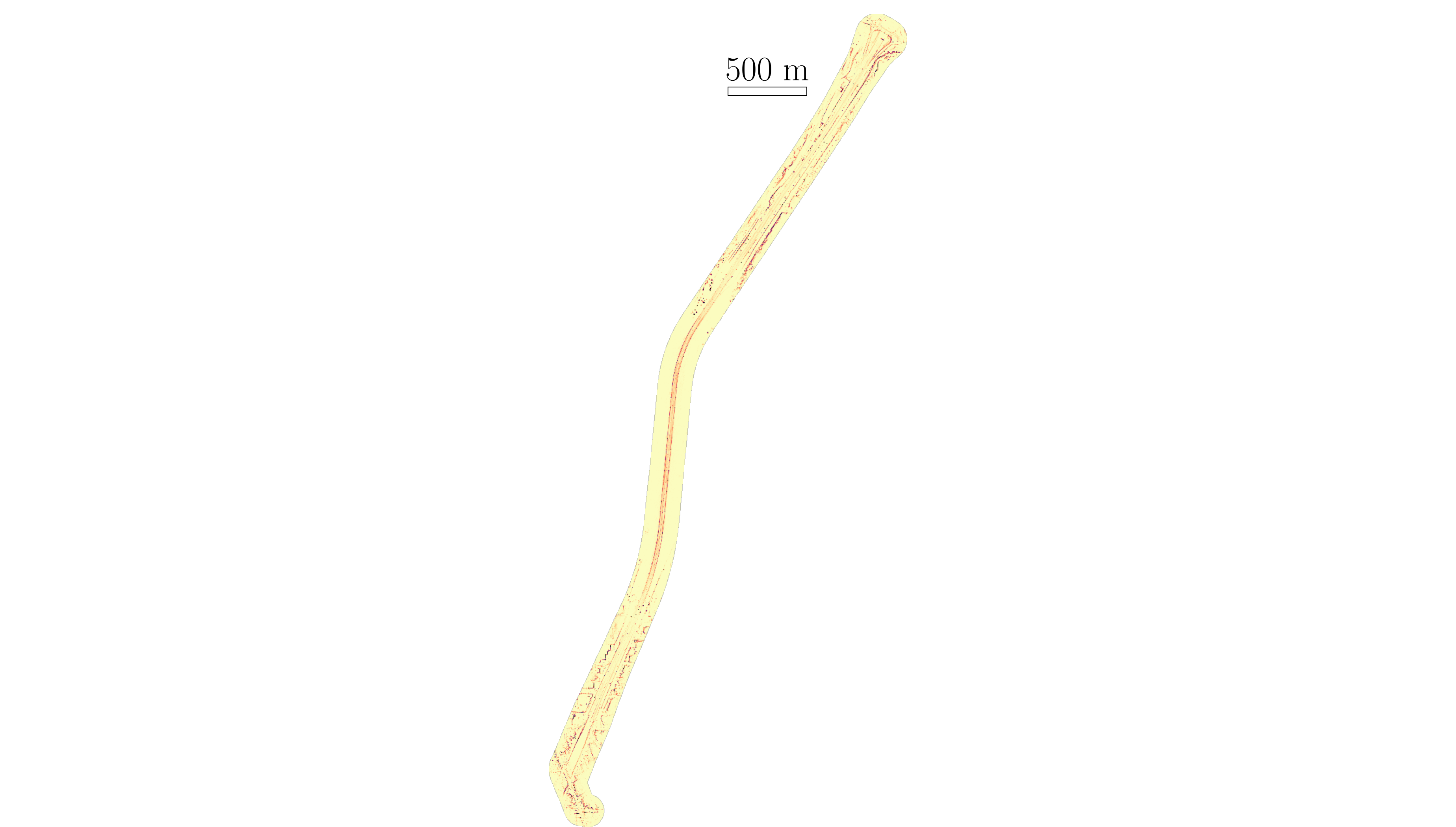}};

        \coordinate (xpos) at ($(map.south west)!0.638!(map.south east)$);
        \coordinate (ypos) at ($(map.south west)!0.887!(map.north west)$);
        \coordinate (zSW)  at ($(xpos |- ypos)$);
        
        \def\zoomSize{2.0cm}
        
        \coordinate (zNE) at ($(zSW) + (\zoomSize,\zoomSize)$);
        
        \draw[dashed, thick] (zSW) rectangle (zNE);

        \node[draw, dashed, thick, inner sep=1pt,
              anchor=north west] (zoomBig)
              at ($(zSW)+(-0.3cm,-3.0cm)$)
        {
          \begin{tikzpicture}
            \clip (0,0) rectangle (6cm,6cm);
            \node[anchor=south west, inner sep=0pt] at (0,0)
              {\includegraphics[
                height=6cm,
                clip,
                trim=0mm 0mm 0mm 0mm
              ]{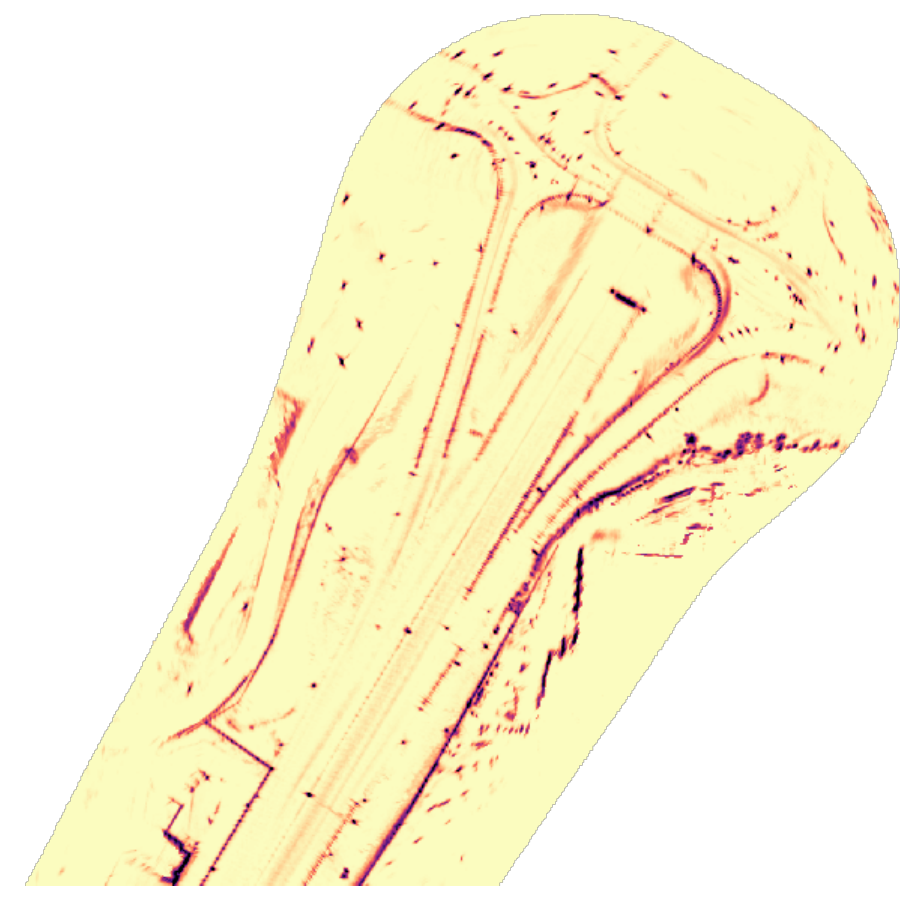}};
          \end{tikzpicture}
        };

        \draw[dashed, thick] (zSW) -- (zoomBig.north west);
        \draw[dashed, thick] (zNE |- zSW) -- (zoomBig.north east);

        \coordinate (xpos2) at ($(map.south west)!0.36!(map.south east)$);
        \coordinate (ypos2) at ($(map.south west)!0.31!(map.north west)$);
        \coordinate (zSW2)  at ($(xpos2 |- ypos2)$);

        \coordinate (zNE2) at ($(zSW2) + (2.3cm,6.3cm)$);
        
        \draw[dashed, thick] (zSW2) rectangle (zNE2);

        \node[draw, dashed, thick, inner sep=1pt,
              anchor=south east] (zoomBig2)
              at ($(zSW2)+(-0.8cm,0.0cm)$)
        {
          \begin{tikzpicture}
            \clip (0,0) rectangle (4.6cm,12.6cm);
            \node[anchor=south west, inner sep=0pt] at (0,0)
              {\includegraphics[
                width=4.6cm,
                clip,
                trim=50mm 0mm 50mm 0mm
              ]{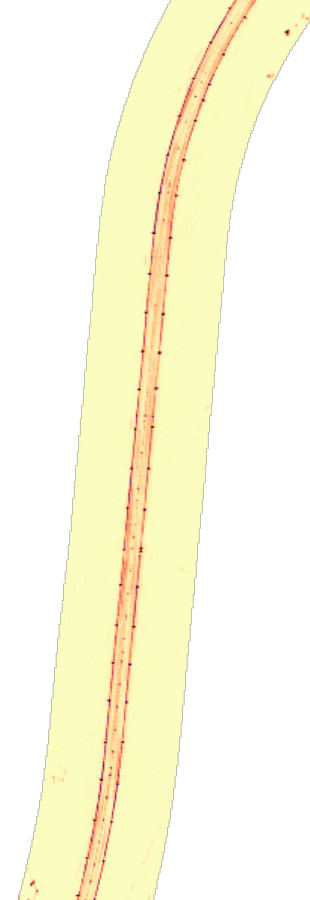}};
          \end{tikzpicture}
        };

        \draw[dashed, thick] (zSW2 |- zNE2) -- (zoomBig2.north east);
        \draw[dashed, thick] (zSW2) -- (zoomBig2.south east);

        \node at (3.5,-16.0) {
          \begin{minipage}{0.4\linewidth}
            \centering
            \includegraphics[width=\linewidth]{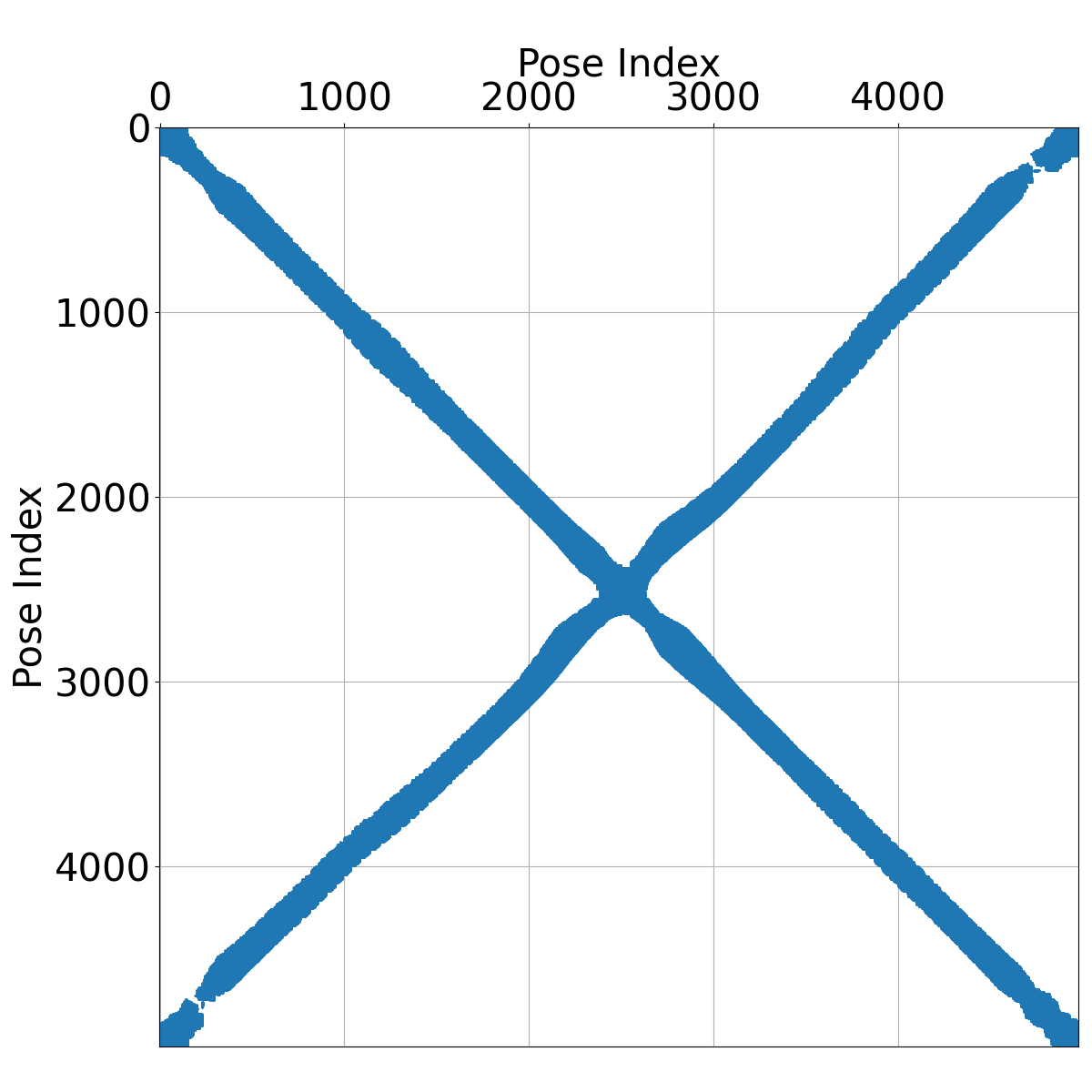}
            \captionof{figure}{Sparsity pattern of $\mbf{H}$ (from \eqref{eq:linear_problem}) for a \texttt{Skyway} sequence.}
          \end{minipage}
        };
        
    \end{tikzpicture}
\end{figure}

\clearpage
\subsection{Forest}

\begin{figure}[htbp]
    \centering
    
    \begin{tikzpicture}
    
        \node[inner sep=0, anchor=north] (map) at (0.0,0.0)
            {\includegraphics[width=\linewidth,
            clip, trim=110mm 0mm 125mm 00mm]{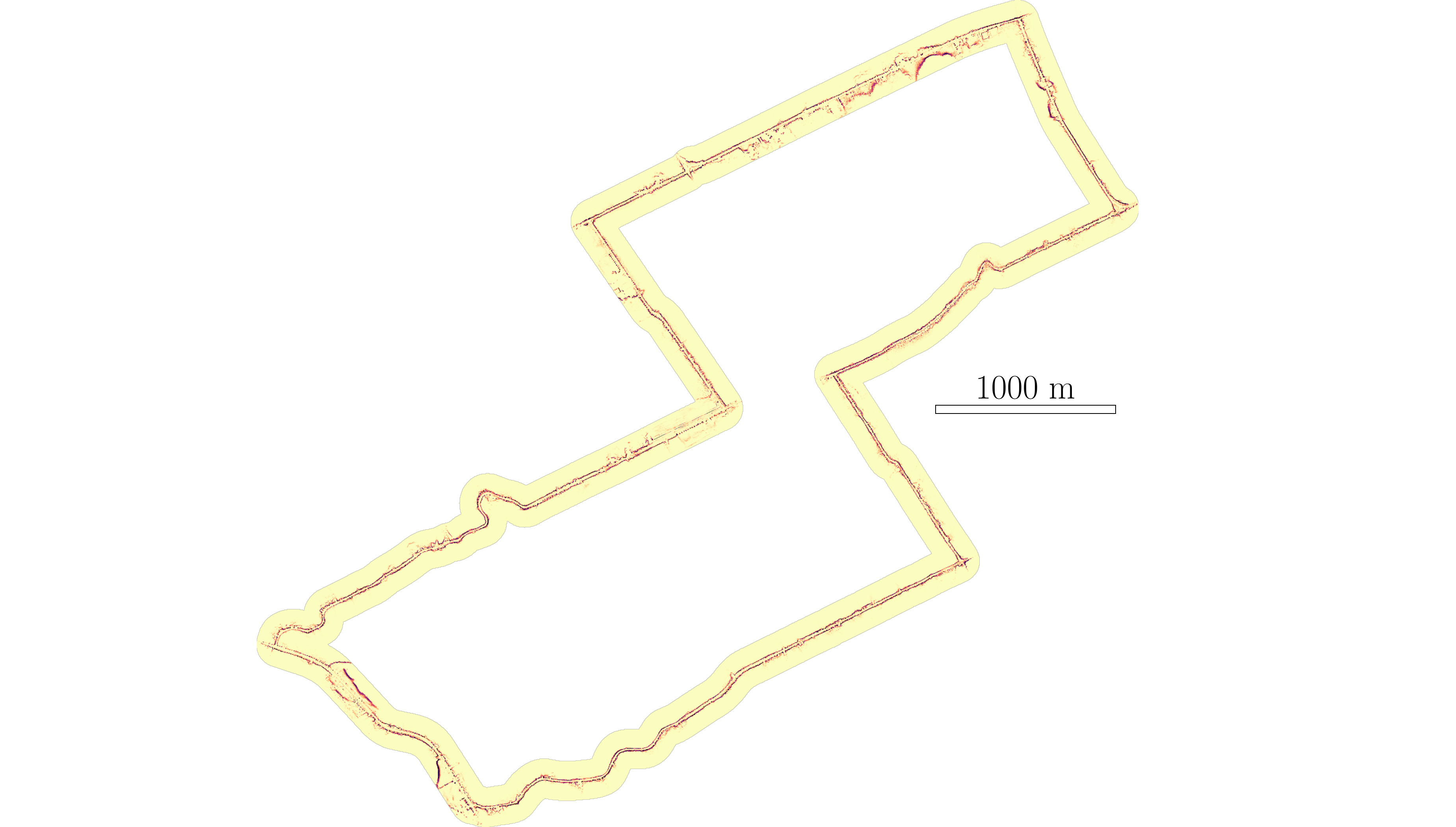}};

        \coordinate (xpos) at ($(map.south west)!0.20!(map.south east)$);
        \coordinate (ypos) at ($(map.south west)!0.31!(map.north west)$);
        \coordinate (zSW)  at ($(xpos |- ypos)$);
        
        \def\zoomSize{2.3cm}
        
        \coordinate (zNE) at ($(zSW) + (\zoomSize,\zoomSize)$);
        
        \draw[dashed, thick] (zSW) rectangle (zNE);

        \node[draw, dashed, thick, inner sep=1pt,
              anchor=north east] (zoomBig)
              at ($(map.north east)+(-12.3cm,-1.1cm)$)
        {
          \begin{tikzpicture}
            \clip (0,0) rectangle (5.5cm,5.5cm);
            \node[anchor=south west, inner sep=0pt] at (0,0)
              {\includegraphics[
                height=5.5cm,
                clip,
                trim=0mm 0mm 0mm 0mm
              ]{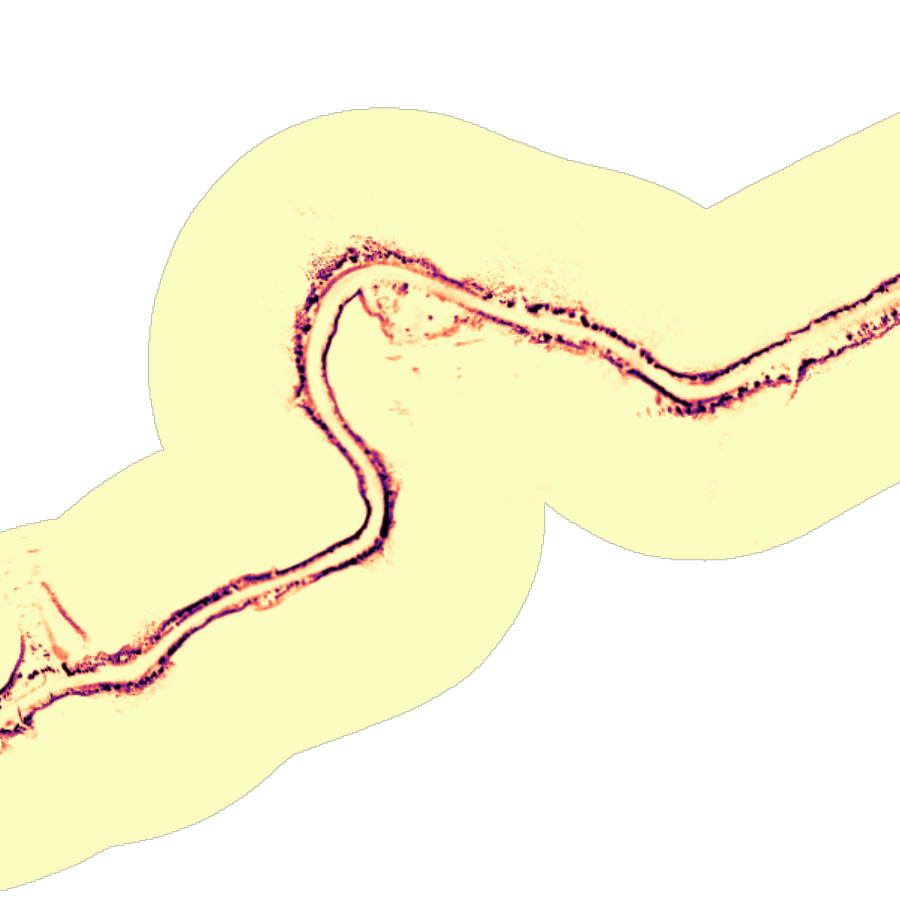}};
          \end{tikzpicture}
        };

        \draw[dashed, thick] (zSW |- zNE) -- (zoomBig.south west);
        \draw[dashed, thick] (zNE |- zNE) -- (zoomBig.south east);

        \node at (5.3,-16.5) {
          \begin{minipage}{0.4\linewidth}
            \centering
            \includegraphics[width=\linewidth]{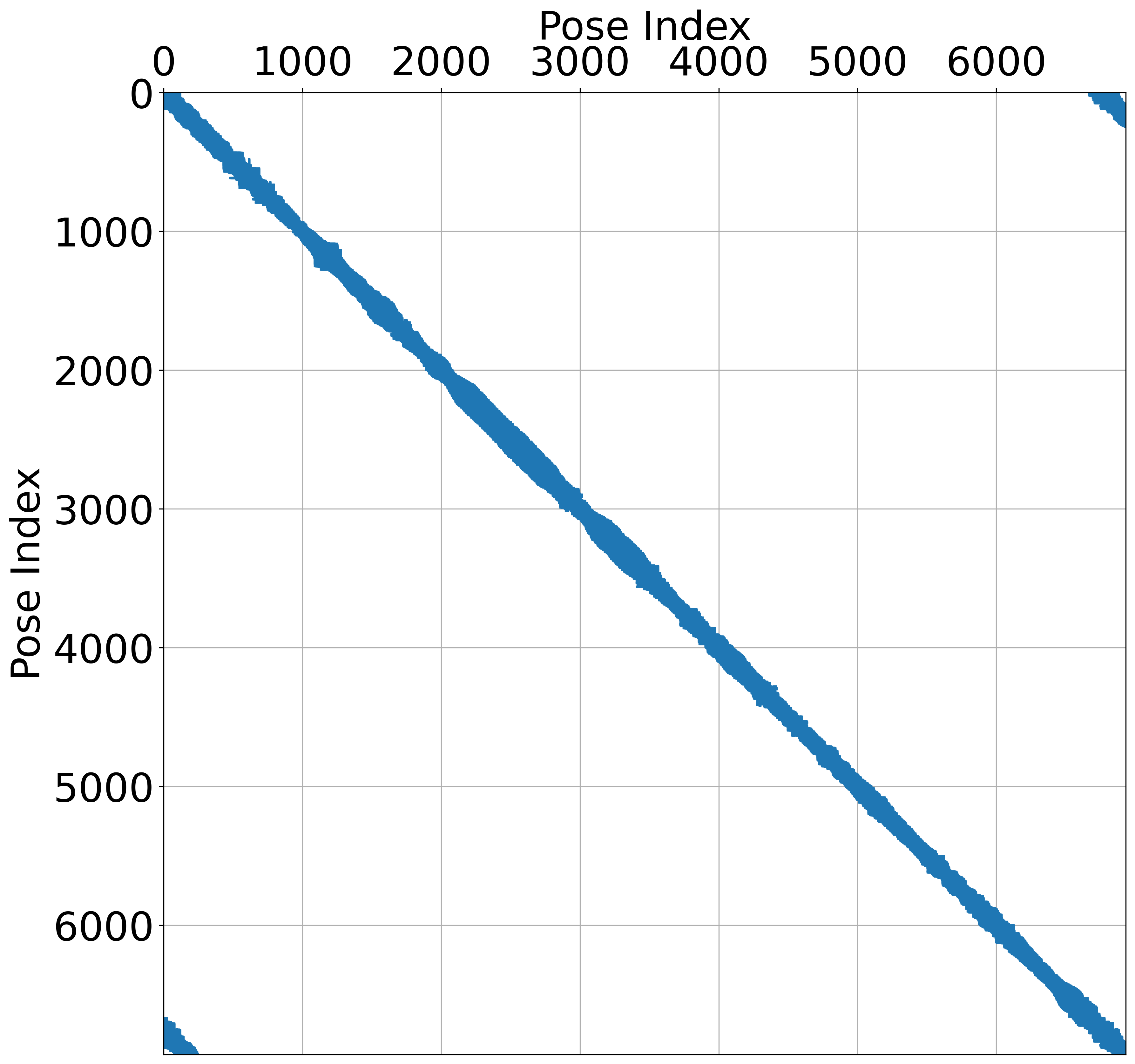}
            \captionof{figure}{Sparsity pattern of $\mbf{H}$ (from \eqref{eq:linear_problem}) for a \texttt{Forest} sequence.}
          \end{minipage}
        };
        
    \end{tikzpicture}
\end{figure}
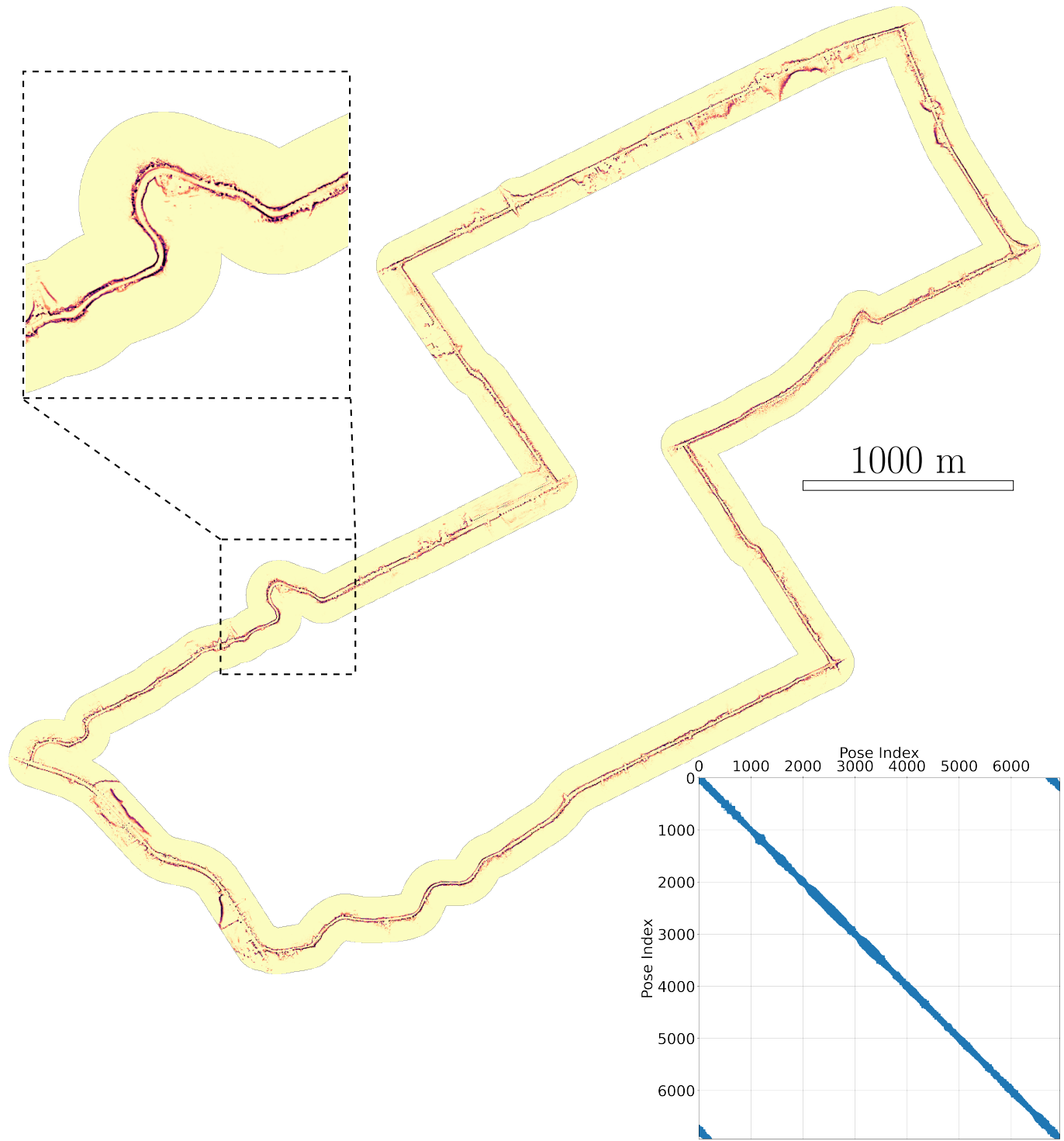

\clearpage
\subsection{Farm}

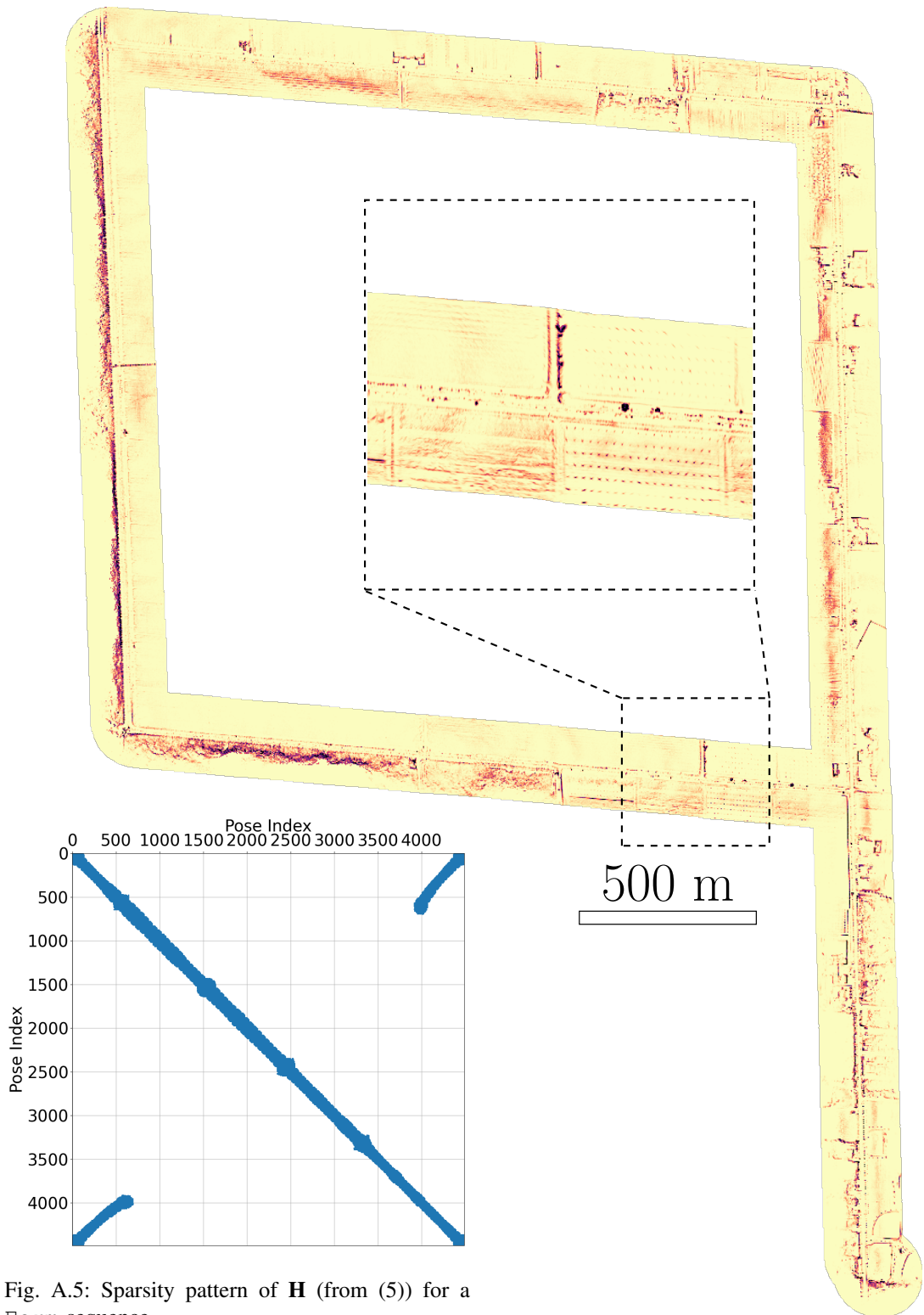
\begin{figure}[htbp]
    \centering
    
    \begin{tikzpicture}
        \node[inner sep=0, anchor=north] (map) at (0.0,0.0)
            {\includegraphics[width=\linewidth,
            clip, trim=155mm 0mm 155mm 00mm]{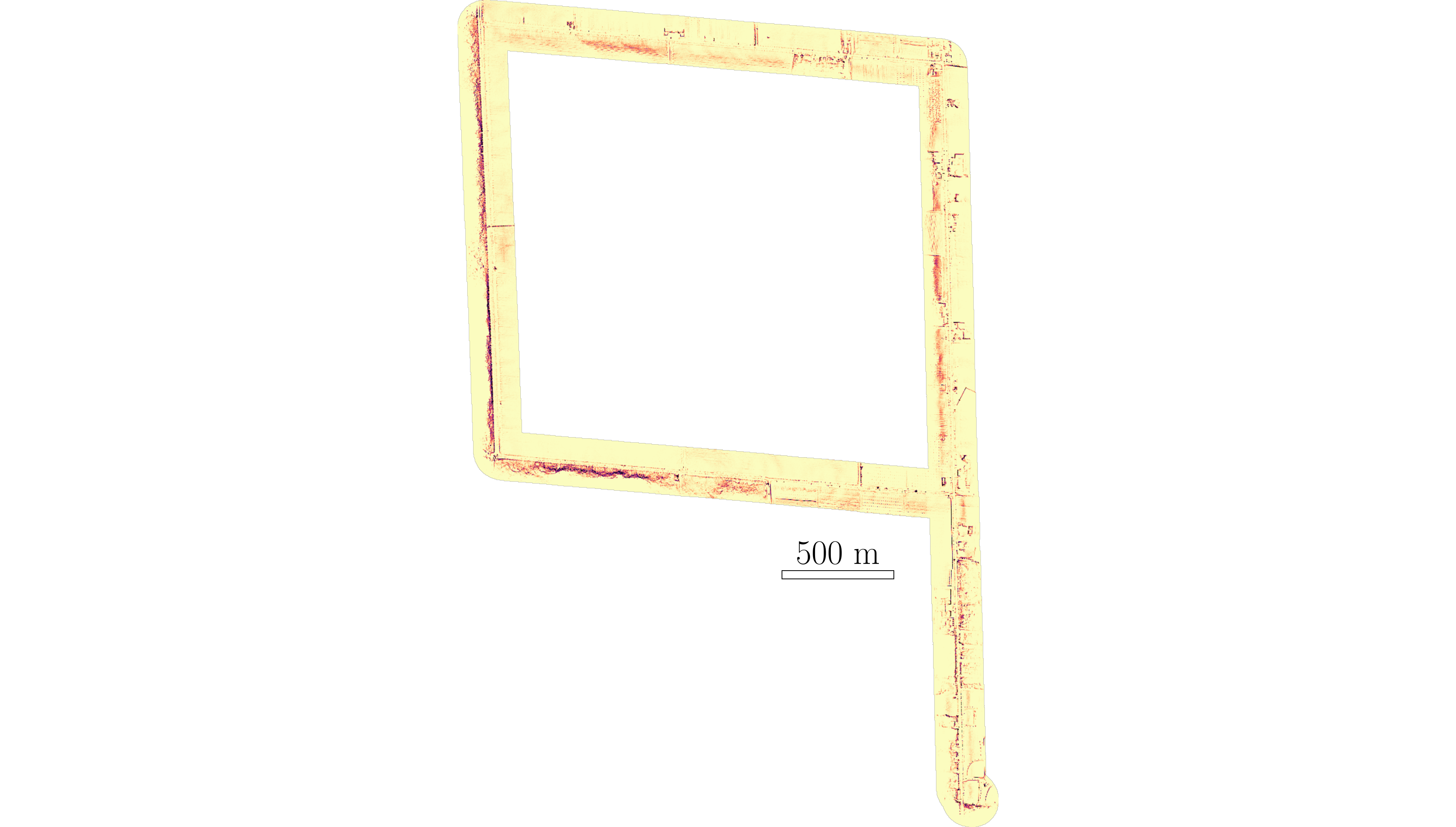}};

        \coordinate (xpos) at ($(map.south west)!0.61!(map.south east)$);
        \coordinate (ypos) at ($(map.south west)!0.36!(map.north west)$);
        \coordinate (zSW)  at ($(xpos |- ypos)$);
        
        \def\zoomSize{2.3cm}
        
        \coordinate (zNE) at ($(zSW) + (\zoomSize,\zoomSize)$);
        
        \draw[dashed, thick] (zSW) rectangle (zNE);

        \node[draw, dashed, thick, inner sep=1pt,
              anchor=north east] (zoomBig)
              at ($(map.north east)+(-5.0cm,-3.cm)$)
        {
          \begin{tikzpicture}
            \clip (0,0) rectangle (6cm,6cm);
            \node[anchor=south west, inner sep=0pt] at (0,0)
              {\includegraphics[
                height=6cm,
                clip,
                trim=0mm 0mm 0mm 0mm
              ]{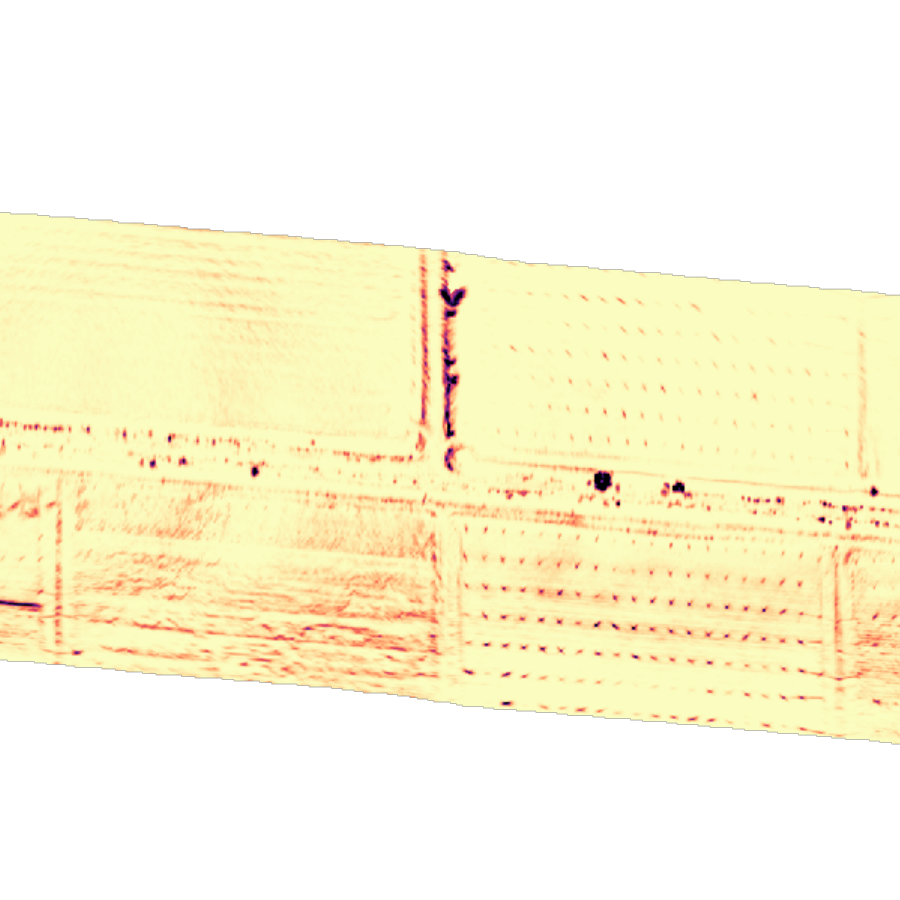}};
          \end{tikzpicture}
        };

        \draw[dashed, thick] (zSW |- zNE) -- (zoomBig.south west);
        \draw[dashed, thick] (zNE |- zNE) -- (zoomBig.south east);

        \node[
          thick,
          inner sep=4pt
        ] at (-4,-16.5) {
          \begin{minipage}{0.4\linewidth}
            \centering
            \includegraphics[width=\linewidth]{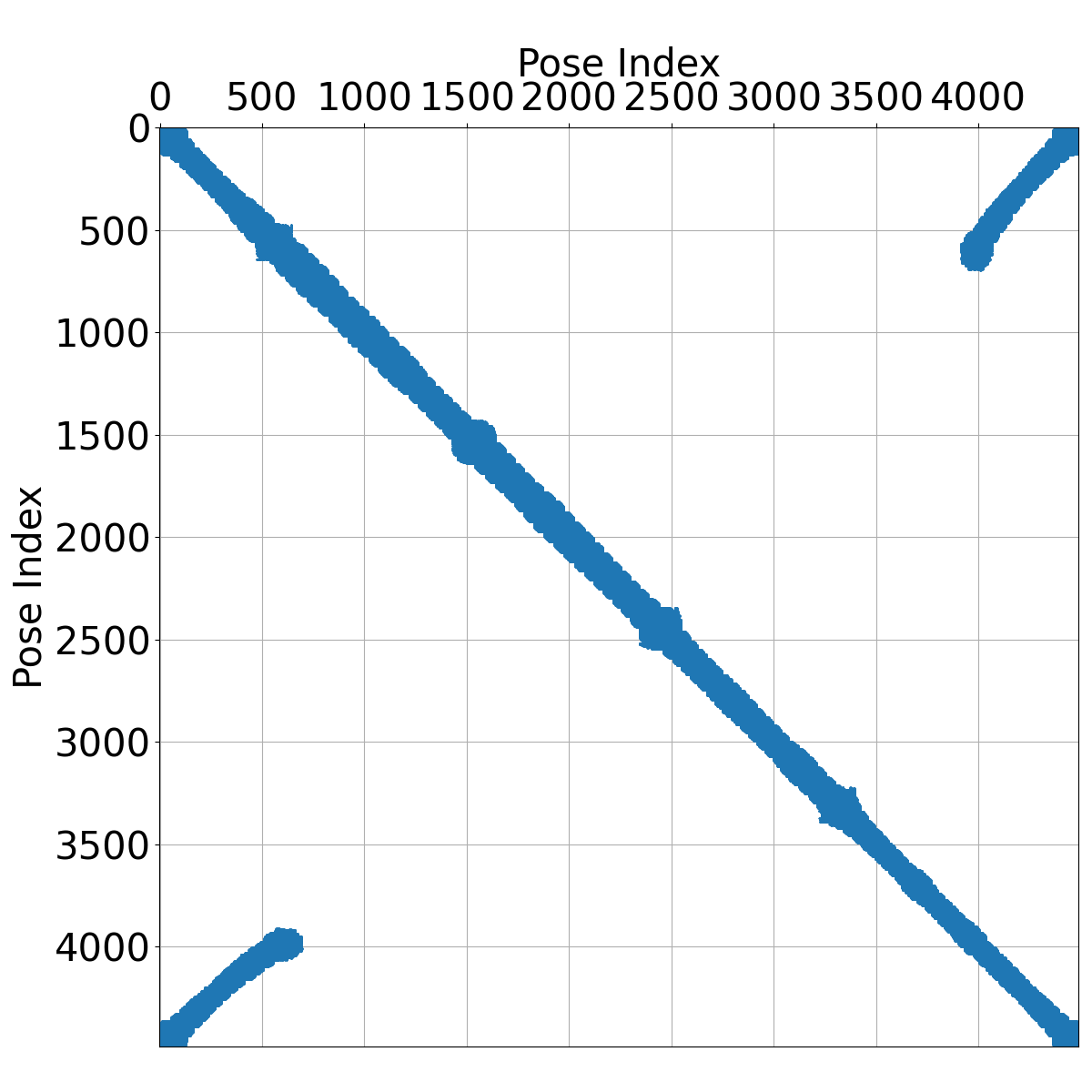}
            \captionof{figure}{Sparsity pattern of $\mbf{H}$ (from \eqref{eq:linear_problem}) for a \texttt{Farm} sequence.}
          \end{minipage}
        };
        
    \end{tikzpicture}
\end{figure}

\end{document}